\pdfoutput=1

\documentclass[11pt]{article}

\usepackage{acl}

\usepackage{times}
\usepackage{latexsym}

\usepackage[T1]{fontenc}

\usepackage[utf8]{inputenc}

\usepackage{microtype}
\usepackage{array} 
\usepackage{inconsolata}

\usepackage{graphicx}
\usepackage{booktabs}
\usepackage{multirow}
\usepackage{longtable}

\usepackage{graphicx}  
\usepackage{subcaption} 
\usepackage{fontawesome5}
\usepackage{mathtools}
\usepackage{paralist}

\usepackage{algorithm}
\usepackage{algorithmic}
\usepackage{amssymb}
\usepackage{xcolor}
\definecolor{YellowOrange}{RGB}{255, 174, 66}  

\usepackage{tcolorbox}

\tcbuselibrary{theorems}
\newtcbtheorem{tldr}{tldr}%
{colback=green!5,colframe=green!35!black,fonttitle=\bfseries}{th}

\author{
  Ziling Cheng$^{1, 2}$\thanks{Equal contribution.} \quad Meng Cao$^{1, 2}$\footnotemark[1] \\
  \quad \textbf{Marc-Antoine Rondeau}$^{1}$ \quad \textbf{Jackie Chi Kit Cheung}$^{1,2,3}$ \\
  $^{1}$ Mila – Quebec Artificial Intelligence Institute \quad \\
  $^{2}$ McGill University \quad
  $^{3}$ Canada CIFAR AI Chair \\
  \texttt{\{ziling.cheng, meng.cao\}@mail.mcgill.ca, \{ma.rondeau, cheungja\}@mila.quebec
}\\
}
\title{Stochastic Chameleons\raisebox{-0.5cm}{\includegraphics[width=1.6cm]{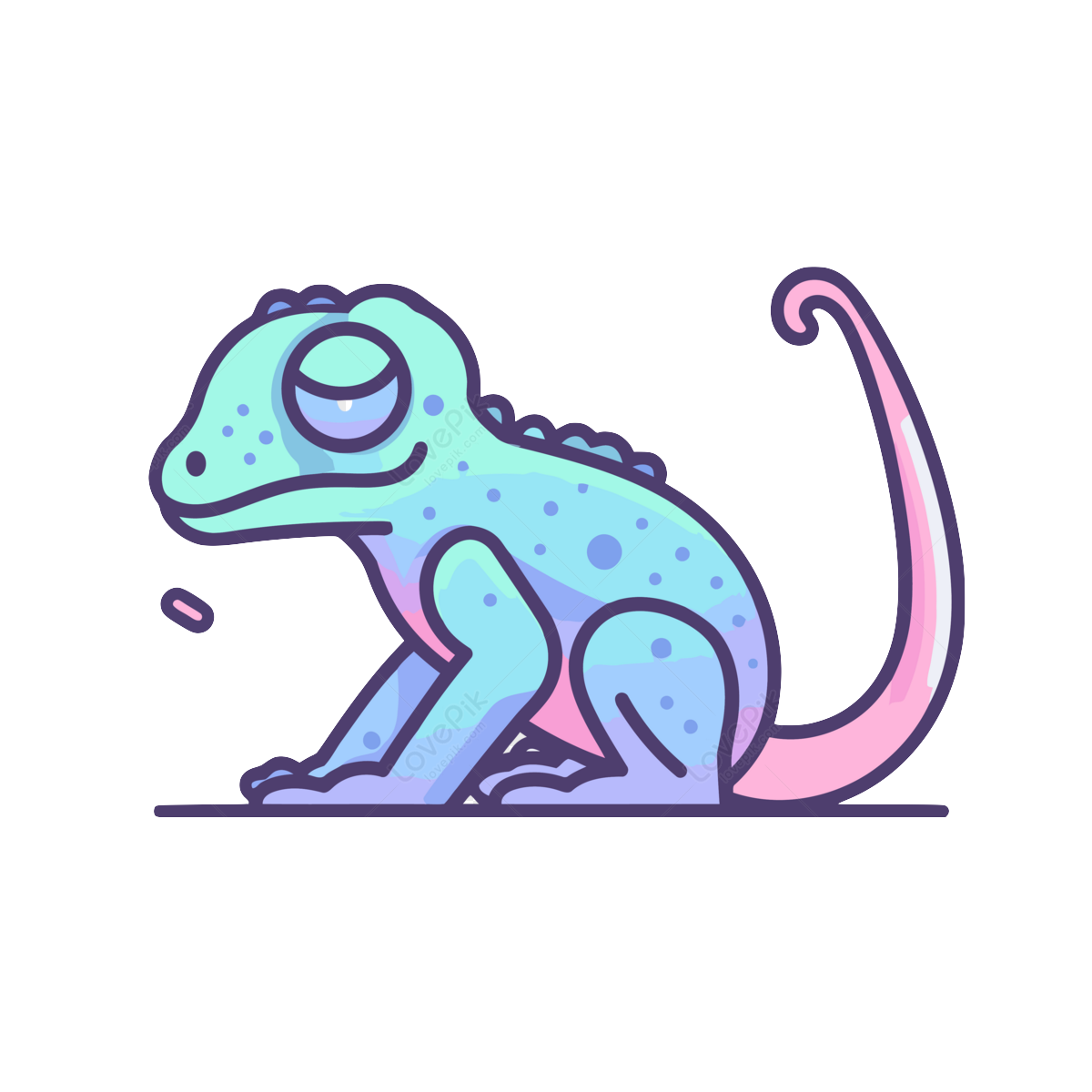}}: Irrelevant Context Hallucinations Reveal Class-Based (Mis)Generalization in LLMs}

\begin{document}
\maketitle
\begin{abstract}
The widespread success of large language models (LLMs) on NLP benchmarks has been accompanied by concerns that LLMs function primarily as stochastic parrots that reproduce texts similar to what they saw during pre-training, often erroneously. But what is the nature of their errors, and do these errors exhibit any regularities? In this work, we examine irrelevant context hallucinations, in which models integrate misleading contextual cues into their predictions. Through behavioral analysis, we show that these errors result from a structured yet flawed mechanism that we term \emph{class-based (mis)generalization}, in which models combine abstract class cues with features extracted from the query or context to derive answers. Furthermore, mechanistic interpretability experiments on Llama-3, Mistral, and Pythia across 39 factual recall relation types reveal that this behavior is reflected in the model's internal computations: (i) abstract class representations are constructed in lower layers before being refined into specific answers in higher layers, (ii) feature selection is governed by two competing circuits --- one prioritizing direct query-based reasoning, the other incorporating contextual cues --- whose relative influences determine the final output. 
Our findings provide a more nuanced perspective on the stochastic parrot argument: through form-based training, LLMs can exhibit generalization leveraging abstractions, albeit in unreliable ways based on contextual cues — what we term \textit{stochastic chameleons}.\footnote{Code available at: \url{https://github.com/ziling-cheng/Irrelevant-Context-Hallucination}.}

\end{abstract}

\section{Introduction}

\begin{figure}
\centering
  \includegraphics[width=\columnwidth]{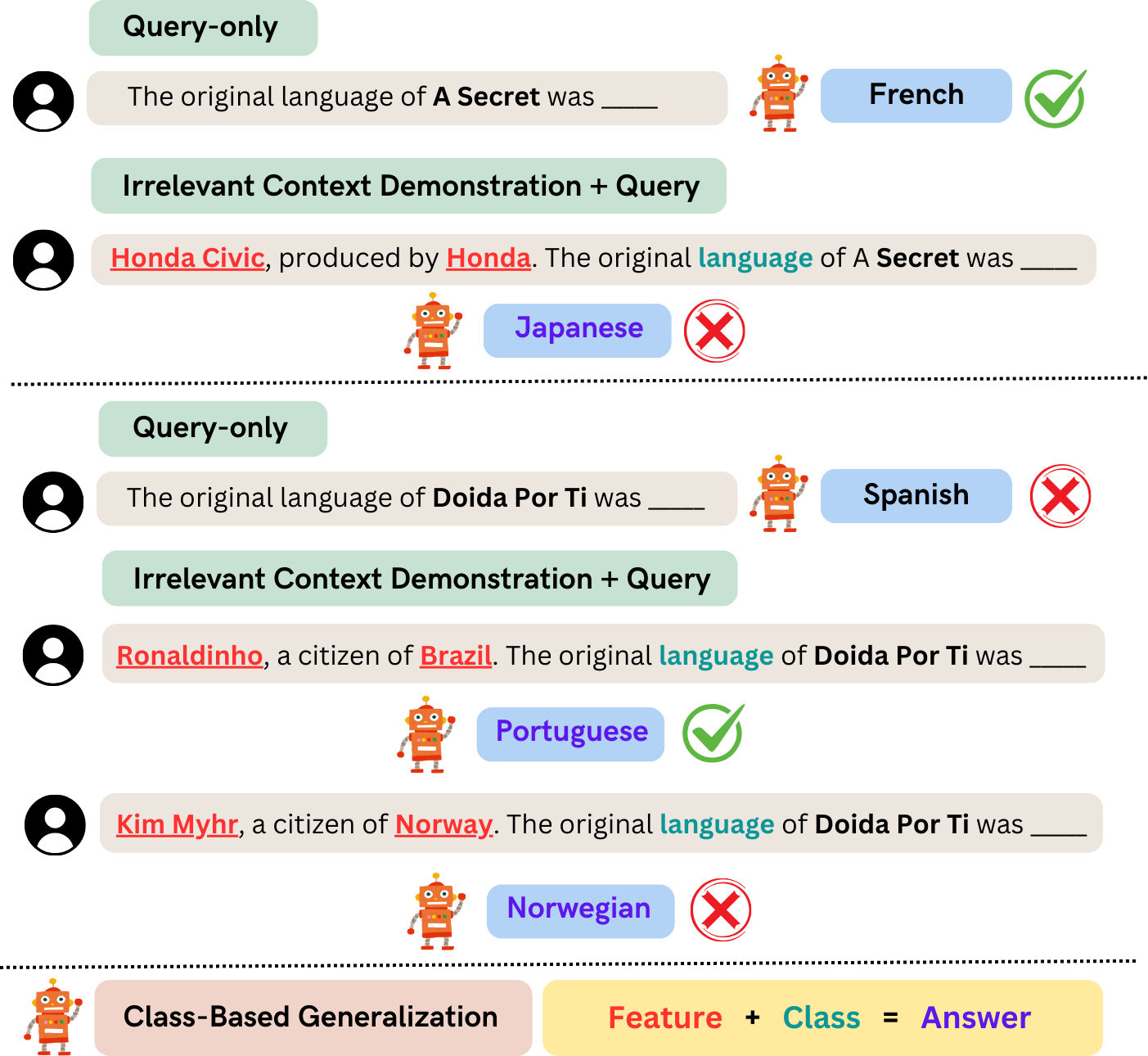}
  \caption{Examples demonstrating class-based (mis)generalization with Llama-3 (8B). \label{Figure:demo_example}}
\end{figure}
The remarkable success of LLMs on various NLP benchmarks has been accompanied by concerns that they function primarily as ``stochastic parrots'' that operate by ``haphazardly stitching together sequences of linguistic forms'' using statistical co-occurrences in pre-training data \cite{bender2021dangers}. 
This view is supported by evidence that LLMs can reproduce training artifacts, exploit spurious correlations, and fail when faced with distribution shifts, among other issues \cite{carlini2021extracting, zhou-etal-2024-explore, dziri2023faith, wu-etal-2024-reasoning, mirzadeh2024gsm}.

In this work, we argue that more deeply examining model errors can reveal insights into LLM behaviors and generalization capabilities. In particular, we examine a specific and underexplored type of error --- \textbf{irrelevant context hallucinations} --- to investigate the mechanisms through which LLMs integrate contextual information into their predictions. We introduce a controlled experimental setting where LLMs receive irrelevant contextual information alongside a query (Figure~\ref{Figure:demo_example}).

By artificially controlling the context and query pairing, this setup allows us to explore how LLMs behave in situations they were unlikely to have encountered in pre-training. Additionally, by focusing on incorrect answers, we sidestep the data leakage issue, which is primarily concerned with memorization of correct answers \cite{balloccu-etal-2024-leak, xu2024benchmarking}. Thus, these controls reduce the possibility that answers stem purely from pattern matching from pre-training data, allowing us to better isolate prediction shifts driven by added context.

Through qualitative analysis of irrelevant context hallucinations, as demonstrated in Figure \ref{Figure:demo_example}, we hypothesize that these errors exhibit structured regularities. We posit that LLMs exhibit a structured but flawed mechanism, which we term the \textbf{class-based (mis)generalization hypothesis}. Specifically, LLMs can leverage abstract class cues (e.g., ``language''), use them to select features in the prompt (e.g., selecting the \textit{country} feature of ``Honda'', instead of the \textit{year} feature)
, and combine these abstract classes with the selected features to produce an answer (e.g., ``Language'' + ``Japan'' $\rightarrow$ ``Japanese''). This hypothesis suggests that LLMs can generalize in a systematic and structured manner in this setting, but as we will show, their reliance on these abstractions is often flawed. In some cases, it leads to correct answers via an incorrect computation (e.g., ``Portuguese'' in Figure \ref{Figure:demo_example}), while in others, it results in hallucinations (e.g., ``Japanese'', ``Norwegian'' in Figure \ref{Figure:demo_example}). 

To validate our hypothesis, we conduct a behavioral analysis of how irrelevant context influences model predictions on Llama-3, Mistral and Pythia. Specifically, we perform annotations on 500 data points and show that 70\% of observed shifts pattern with our class-based generalization hypothesis. Moreover, statistical analyses confirm that this phenomenon is systematic rather than due to chance or being query-dependent.

We provide further evidence of this generalization mechanism via mechanistic interpretability experiments which probe the model’s internal computations across Transformer layers \cite{NIPS2017_3f5ee243}. Our findings reveal two key mechanisms that further support our hypothesis:
(i) LLMs make hierarchical class-to-instance predictions; i.e., they construct abstract class representations (e.g., ``languages'') before refining them to more specific answers (e.g., ``Japanese''). (ii) Feature selection is governed by competing circuits: we identify one pathway that prioritizes direct query-based reasoning and another that incorporates contextual cues. Their relative strength determines the final output. Attention knockout experiments show that ablating key heads involved in the context-based pathway can flip model predictions (e.g., flipping ``Japanese'' to ``French''), further confirming this competitive interaction. These findings support the class component of our hypothesis and illustrate how models select features to combine with the abstract class.

Crucially, our findings suggest that LLMs go beyond mere parroting: they exhibit a form of generalization that leverages abstract class structures based on contextual cues in ways that are systematic, though not necessarily reliable. These abstractions result from next-token prediction during pre-training and extend beyond simple ontological hierarchies (e.g., superset-subset relationships), shaping the model’s internal feature selection process. To capture this behavior, we propose the metaphor of \emph{stochastic chameleons} — models that, like a chameleon changing colors in response to environmental and internal signals, dynamically shift their outputs based on contextual cues. However, this does not contradict the central claim of the stochastic parrot argument: that LLMs lack true language understanding when trained solely on linguistic form \cite{bender-koller-2020}.

In summary, our main contributions are:
\begin{compactitem}
    \item We introduce a novel setting that isolates how LLMs integrate irrelevant contexts, distinguishing generalization from memorization. 
    \item We provide empirical evidence that LLMs exhibit class-based (mis)generalization, demonstrating sensitivity to abstract class structures beyond statistical co-occurrences.
    \item We uncover the internal computational mechanisms of class-based generalization, revealing competing circuits and hierarchical class representations.
    \item We propose a behavioral analysis framework that moves beyond accuracy-based evaluation, emphasizing the importance of understanding LLMs' internal mechanisms.
\end{compactitem}

\begin{figure*}[h!]
\centering
  \includegraphics[width=2\columnwidth]{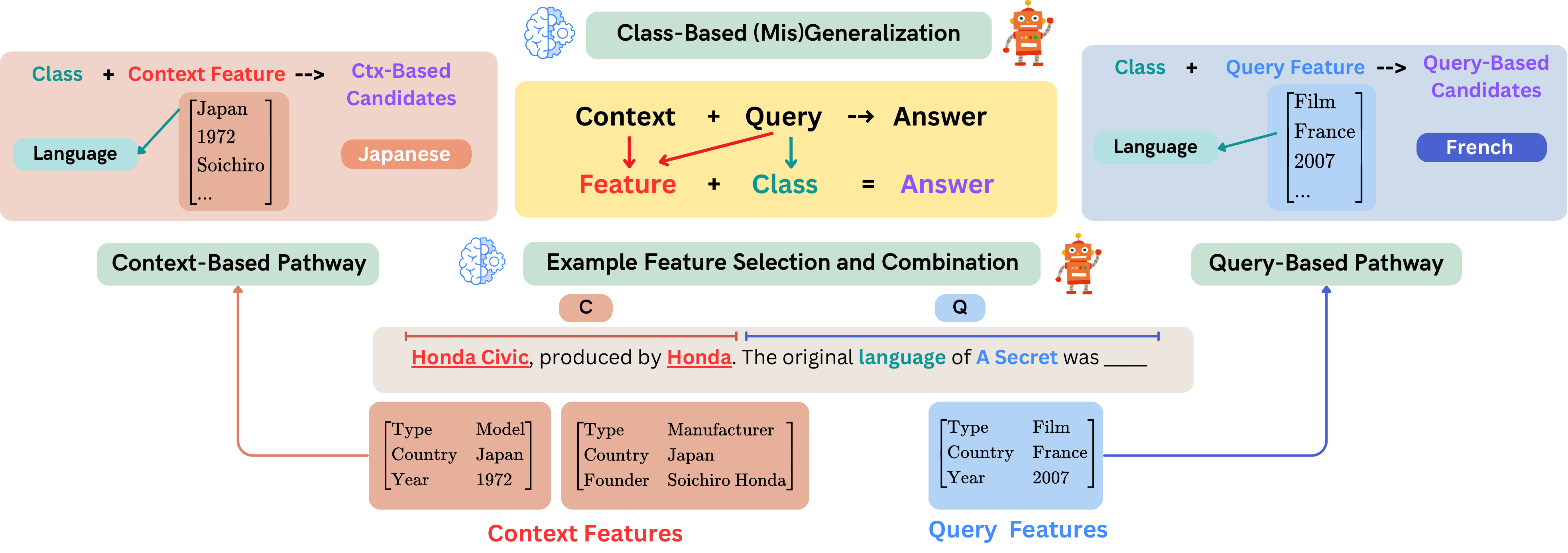}
  \caption{Class-based generalization framework: feature selection and combination. \label{Figure:framework}}
\end{figure*}

\section{Related Work}


\paragraph{LLM Evaluation} 
Traditional NLP evaluation prioritizes test set performance but often overlooks how models arrive at their final answers. For LLMs trained on Internet-scale data, distinguishing genuine generalization from memorization or spurious correlations is challenging, especially with potential data leakage \cite{dziri2023faith, wu-etal-2024-reasoning, zhou-etal-2024-explore, balloccu-etal-2024-leak, xu2024benchmarking}. Prior work addresses this through data extraction \cite{carlini2021extracting}, statistical control \cite{min-etal-2022-rethinking}, adversarial perturbations \cite{mirzadeh2024gsm}, and error analysis \cite{dziri2023faith}. In contrast, we take a behavior-focused approach in a controlled setting, reducing the likelihood of pure pattern matching. Additionally, by analyzing errors — often ignored by accuracy-based metrics — we gain deeper insights into model mechanisms.

\paragraph{Irrelevant Context Hallucinations} 
Hallucinations in text generation have been studied in the absence of context \cite{mckenna2023sources, kang2024unfamiliar, meng2022locating} and in cases with relevant context \cite{cao-etal-2020-factual, cao-etal-2022-hallucinated, maynez-etal-2020-faithfulness, lee2018hallucinations, adlakha-etal-2024-evaluating, chuang-etal-2024-lookback, petroni2020how, li-etal-2023-large}. We focus on irrelevant context hallucinations, where extraneous context influences predictions. Unlike prior work on evaluating or mitigating such errors \cite{yoran2024making, Cuconasu_2024, wu2024how, petroni2020how, cao-etal-2022-learning,  li-etal-2023-large, shi2023large, mirzadeh2024gsm}, we explain their underlying class-based (mis)generalization mechanisms, conceptually and mechanistically.

\paragraph{Interplay Between Contextual and Parametric Knowledge} Prior work primarily focuses on resolving conflicts between relevant contexts and parametric knowledge -- determining when models should rely on one over the other \cite{jin-etal-2024-cutting, xu-etal-2024-knowledge-conflicts, su2024textttconflictbank, yuan-etal-2024-discerning, marjanovic-etal-2024-dynamicqa, neeman-etal-2023-disentqa, chen-etal-2022-rich, longpre-etal-2021-entity}. In contrast, we study irrelevant context hallucinations, where unrelated and non-contradicting context still shapes predictions, competing with parametric knowledge related to the query.

\paragraph{Mechanistic Interpretability} 
Mechanistic interpretability methods \cite{Olah2022, Neel2023} reverse-engineer LLMs via vocabulary projection \cite{belrose2023eliciting, geva-etal-2022-transformer, nostalgebraist2020} and computational interventions \cite{ghandeharioun2024patchscopes, stolfo-etal-2023-mechanistic, finlayson-etal-2021-causal, hong2025reasoning, cheng2025llmsreasonabstractlymath}. Extending prior work \cite{merullo-etal-2024-language, wu2024retrieval, lv2024interpreting, yu-etal-2023-characterizing, yu-etal-2024-mechanistic, geva-etal-2022-transformer}, we use these techniques to uncover how LLMs are influenced by irrelevant contexts. By linking behavioral analysis with internal mechanisms, we provide a mechanistic perspective on irrelevant context hallucinations.

\begin{table*}[h!]
\centering
\small
\begin{tabular}{@{}m{6cm}m{1cm}m{1.5cm}m{1.8cm}m{1.5cm}m{2cm}@{}}
\toprule
\textbf{$C+Q$} & \textbf{$A_{Q}$}  & $C_{\text{cand.}}$ & $Q_{\text{cand.}}$ & \textbf{$A_{C+Q}$}  & \textbf{Case}\\
\midrule
\textbf{C}: Honda Civic, produced by Honda. \newline 
\textbf{Q}: The original language of A Secret was
& French 
& \textbf{Japanese}
& French, \newline English
& \textbf{Japanese} 
& Context-dominant\\
\midrule
\textbf{C}: City of Boroondara is in Melbourne. \newline 
\textbf{Q}: Prime Minister of Malaysia is a legal term in
& Malaysia 
& Australia
& \textbf{Malaysia}, Malaysian
& \textbf{Malaysia} 
& Query-dominant\\
\bottomrule
\end{tabular}
\caption{Examples of context- and query-dominant categorizations with context- and query-based candidates.}
\label{tab:dominant_examples}
\end{table*}

\section{Framework and Hypothesis}\label{sec:framework}
In this section, we describe the abstract framework illustrated in Figure \ref{Figure:framework} and present our class-based (mis)generalization hypothesis.

\paragraph{Setting}  
Consider a query \( Q \) representing the question of interest and an irrelevant context \( C \) prepended to it. Let \( A_Q \) denote the model’s answer to \( Q \) alone and \( A_{C+Q} \) denote the answer with the added context (contextual answer). We define \textbf{context features} and \textbf{query features} as sets of properties or attributes of the entities in \( C \) and \( Q \), respectively. For example, for context features in Figure~\ref{Figure:framework}, the feature names are \{\textit{Type, Country, Year, etc.}\}, with corresponding feature values \{\textit{Model, Japan, 1972, etc.}\}. The \textbf{class} of \( A_{C+Q} \) derived from $Q$ (e.g., ``languages'') determines which features the model should prioritize (e.g., derive ``Japanese'' based on the ``Country: Japan'' feature).

\paragraph{Class-Based (Mis)Generalization Hypothesis}  
Given the setup \( C + Q \), when the context influences the model predictions, instead of relying solely on the query, we hypothesize a structured mechanism by which the model integrates contextual information into its predictions. Specifically, we propose \textbf{class-based generalization}, where language models process context in two steps: they first \textit{derive} an abstract class (e.g., ``languages'') and then \textit{select} and \textit{combine} relevant features from \( C \) or \( Q \) (e.g., ``Japan'' or ``France'') to generate an answer (e.g., ``Japanese' or ``French'').

Let \textbf{query-based candidates} be answers derived from query features combined with the class (e.g., ``French'') and \textbf{context-based candidates} be answers derived from context features (e.g., ``Japanese'').  If \( A_{C+Q} \) is the query-based candidate, we define the case as \textbf{query-dominant}; otherwise, it is \textbf{context-dominant}. These terms emphasize the final outcome rather than the intermediate steps. See Table~\ref{tab:dominant_examples} for examples.

A special case arises when a token of the expected class is already present in the prompt (Appendix \ref{appendix: hypothesis}), making the model more likely to copy it directly \cite{jiang2024llms}.

\section{Dataset and Experimental Design}\label{sec:dataset}
\paragraph{Models \& Datasets} We evaluate three pretrained LM families — Llama-3 (8B, 70B) \cite{llama3modelcard}, Mistral v0.3 (7B) \cite{jiang2023mistral}, and Pythia (6.9B-deduped, 12B-deduped) \cite{biderman2023pythia} — using their base versions to assess raw model behavior. We use the ParaRel dataset \cite{elazar-etal-2021-measuring}, which consists of 39 factual QA subdatasets. Dataset statistics are provided in Table~\ref{tab: pararel_dataset} in the Appendix \ref{appendix: dataset}. Both \( Q \) and \( C \) are sourced from these datasets. Experiments are run on two RTX8000 GPUs.

\paragraph{Experimental Setup}  
We compare two conditions:  1) \textbf{Q-only}, where each \( Q \) is formatted using a predefined template and a subject-relation-object $(s, r, o)$ triplet from ParaRel, resulting in 27.6K queries. The model’s top-1 prediction is \( A_Q \). 2) \textbf{C+Q}, where each \( Q \) is prepended with context demonstrations from other subdatasets spanning various relation types, introducing controlled contextual variation. We randomly sample 100 examples per subdatasets\footnote{P264 has only 53 examples, so we include all of them.}, generating 3,900 context variations per query, totaling 106M examples. The model’s top-1 prediction is \( A_{C+Q} \).  

\paragraph{Context- and Query-Based Candidates} 
To make the definitions from Sec.~\ref{sec:framework} precise, 
we define a \textbf{context-based candidate} \( x \in  C_{\text{cand.}}\) to be a candidate among the top three\footnote{A threshold of three ensures that classified context-based candidates are strongly influenced by context.} predictions under \( C+Q \) but not among the top ten\footnote{A threshold of 10 ensures that classified context-based candidates are not plausible answers under $Q$ alone.} predictions under $Q$. A \textbf{query-based candidate} \( x \in  Q_{\text{cand.}}\) appears in the top predictions under both conditions. Note that $A_{C+Q}^{\text{top3}} = C_{\text{cand.}} \cup Q_{\text{cand.}}$. See Table~\ref{tab:dominant_examples} for examples.

\vspace{-10pt}
\begingroup
\small
\begin{align}
    A_{C+Q}^{\text{top3}} &:= \{x \mid x \in \text{top 3 candidates under } C+Q\} \label{set:X} \\
    A_{Q}^{\text{top10}} &:= \{x \mid x \in \text{top 10 candidates under } Q\} \label{set:Y} \\
    C_{\text{cand.}} &:= \{x \mid x \in A_{C+Q}^{\text{top3}} \text{ and } x \notin A_{Q}^{\text{top10}}\} \\
    Q_{\text{cand.}} &:= \{x \mid x \in A_{C+Q}^{\text{top3}} \text{ and } x \in A_{Q}^{\text{top10}}\}
\end{align}
\endgroup

\section{Behavioral Analysis of Contextual Answers}

We now investigate how irrelevant context influences model predictions, verifying our class-based (mis)generalization hypothesis through textual-level behavioral analysis. Specifically, we examine: (1) whether irrelevant context causes behavioral changes (Sec. \ref{sec: behaviour_changes}), (2) whether the influence of irrelevant context aligns with our hypothesis (Sec. \ref{sec: annotation}). (3) whether the observed correlation between irrelevant context and context-based candidates is statistically significant (Sec. \ref{sec: testing}).

\begin{table}[h]
    \centering
    \small
    \begin{tabular}{@{}p{1cm}p{3cm}p{0.6cm}p{0.6cm}p{0.8cm}@{}}
        \toprule
        \textbf{Case} & \textbf{Top-3 Candidates} & \textbf{Llama} & \textbf{Mistral} & \textbf{Pythia} \\
        \midrule
        No influence & 1. All query-based ($C_{\text{cand.}}= \emptyset$)  & 47.9\% & 48.0\% & 39.3\%\\
        \midrule
        \multirow{2}{=}{Query-dominant}  & 2. Mix, top-1 is query-based & 27.9\% & 25.7\% & 27.2\%\\
        \midrule
        \multirow{4}{=}{Context-dominant} 
        & 3. Mix, top-1 is context-based & 15.1\% & 17.0\% & 19.2\%\\
        \cmidrule{2-5}
        & 4. All context-based ($Q_{\text{cand.}}= \emptyset$) & 10.1\% & 10.3\% & 14.3\%\\
        \bottomrule
    \end{tabular}
    \caption{Breakdown of samples according to the composition of $A_{C+Q}^{\text{top-3}}$, based on 106M datapoints. Detailed results can be found in Table \ref{tab:model_comparison_detailed} in the Appendix.}
    \label{tab:model_comparison}
\end{table}

\begin{table*}[h!]
\centering
\small
\begin{tabular}{@{}p{1.15cm}p{1.1cm}p{13cm}@{}}
\toprule
\textbf{Query Type} & \textbf{Ctx. Type} & \textbf{Context Demonstration + Query and Answer} \\
\midrule
\multirow{4}{*}{Language} 
& Person/\newline Music 
& 
\textbf{Prompt:} {\color{blue}Amilcare Ponchielli} plays {\color{blue}opera}. The original language of A Hunting Accident was \_\_\_\_\_\_ \newline
\textbf{ Answer:}  $A_{C+Q}=$ {\color{blue}Italian}, $A_{Q}=$ English \\ 

\cmidrule(l){2-3}
& Make/\newline Model 
& 
\textbf{Prompt:} {\color{blue}Toyota Alphard}, produced by {\color{blue}Toyota}. The original language of A Hunting Accident was \_\_\_\_\_\_
\textbf{ Answer:} $A_{C+Q}=$ {\color{blue}Japanese}, $A_{Q}=$ English \\ 
\midrule
\multirow{4}{*}{Place} 
& Person/\newline Religion 
& 
\textbf{Prompt:} {\color{blue}Indo-Greek Kingdom} is follower of {\color{blue}Buddhism}. Alpha Island is a part of the continent of \_\_\_\_\_\_
\textbf{ Answer:} $A_{C+Q}=$ {\color{blue}Asia}, $A_{Q}=$ Alpha \\ 
\cmidrule(l){2-3}
& Place
& 
\textbf{Prompt:} {\color{blue}Council of States of Switzerland} is a legal term in {\color{blue}Switzerland}. Alpha Island is a part of the continent of \_\_\_\_\_\_
\textbf{ Answer:} $A_{C+Q}=$ {\color{blue}Europe}, $A_{Q}=$ Alpha \\
\bottomrule

\end{tabular}
\caption{Examples of context-based candidates across different query and context types.}
\label{tab:case-study-non-copied}
\end{table*}

\subsection{Behavioral Changes Induced by Irrelevant Context}\label{sec: behaviour_changes}
In this section, we investigate whether adding irrelevant context leads to behavioral changes in model predictions. From an accuracy perspective, we observe a slight decrease: across 39 subdatasets, the accuracy for Llama-3 drops from 47.2\% to 43.1\%, and for Mistral, from 38.2\% to 35.3\% (Table \ref{tab:contextual_hallucinations_stats_other_model} in Appendix). While these changes are modest, accuracy alone does not provide a complete picture of changes in model predictions. To address this gap, we measure the answer change rate ($\Delta$ Rate) after adding the irrelevant context: $\Delta \text{Rate} = \frac{|A_{C+Q}\neq A_Q|}{\# \text{ datapoints}}$. For Llama-3, 38.3\% of responses changed after adding irrelevant context, while for Mistral, nearly half of the datapoints (48.0\%) experience a shift in predictions (Table \ref{tab:contextual_hallucinations_stats_other_model} in Appendix).

We further examine the cases under the \(C + Q \) condition based on the composition of ($A_{C+Q}^{\text{top-3}}$) (Table~\ref{tab:model_comparison}). Roughly 48\%\footnote{Due to the conservative choice of 10 for $A_Q^{\text{top-10}}$, some answers in case 1 might also be context-based but already appear in the top-10 predictions under the $Q$ condition. Therefore, we exclude these cases from further analyses.} of samples are unaffected by the irrelevant context for Llama and Mistral (case 1), meaning all top-3 candidates are query-based). However, when predictions are influenced by the added context (cases 2, 3 and 4), about half of these instances (49.5\% for Llama, 52.5\% for Mistral) become context-dominant. These results demonstrate the influence of irrelevant contexts, even if the overall accuracy is little changed.

\subsection{Human Annotation of Context-Based Candidates}\label{sec: annotation}
Next, we examine whether these behavioral changes pattern with our class-based generalization hypothesis. To do so, we annotate context-based candidates, which capture the shifts induced by irrelevant context. We assess whether each answer explicitly integrates \textit{identifiable features} from the context \textit{and} combines them with the \textit{expected class} indicated by the query. Annotation procedure and examples are provided in Appendix~\ref{appendix: annotation}.

We perform this annotation on a randomly sampled set of 500 context-based candidates across different subdatasets. Our results reveal that 81.6\% of the responses incorporate features from the provided context, 84.4\% belong to the correct class, and 71.0\% satisfy both criteria -- combining identifiable context features with the correct abstract class. This finding provides strong evidence for our hypothesis as a majority of these samples can be explained by the hypothesis. Table~\ref{tab:case-study-non-copied} provides illustrative examples of the model’s output adapting to contextual cues.

\subsection{Statistical Validation of Contextual Influence}\label{sec: testing}
Next, we investigate whether the correlation between irrelevant context and context-based candidates is statistically significant. To quantify the dependence between a context $C$ (e.g., \textit{Honda}) and its associated context-based candidate $C_{\text{cand.}}$ (e.g., \textit{Japanese}), we compute the pointwise mutual information (PMI) between them. Specifically, we sample 100 distinct contexts from various subdatasets. Each context is paired with 100 different queries belonging to the same expected class (e.g., \textit{languages}, \textit{places}, etc.), resulting in 10,000 instances per class. Since context-based candidates are determined independently of the queries, each context $C_i$ is paired with its corresponding candidate $C_{\text{cand.,i}}$, regardless of the 100 queries. This yields 100 pairs of $(C_i, C_{\text{cand.}_,i})$ per class, such as (\textit{Honda}, \textit{Japanese}) for languages, and (\textit{Honda}, \textit{Japan}) for places. The mean PMI across the 100 pairs of each class is computed as:
\begin{align}\mu_{\text{observed}} = \frac{1}{100} \sum_{i=1}^{100} \text{PMI}(C_i, C_{\text{cand.}_,i}) \\
\text{PMI}(C_i, C_{\text{cand.}_,i}) = \log \frac{P(C_i, C_{\text{cand.}_,i})}{P(C_i) P(C_{\text{cand.}_,i})}.
\label{eq:pmi}
\end{align}
In this formula, $P(C_i) = 1/100$, since we have 100 distinct contexts. $P(C_{\text{cand.}_,i})$ is estimated based on its frequency among all 10,000 generated answers $A_{C+Q}$ for the given expected class. Similarly, $P(C_i, C_{\text{cand.}_,i})$  is computed from its co-occurrence within these samples. Across all models and expected classes, the mean PMI is approximately 4, suggesting a strong association between contexts and their corresponding candidates. To formally assess statistical dependence, we perform a one-sample t-test against the null hypothesis $\mathbb{E}[\text{PMI}(C_i, C_{\text{cand.}_,i})] = 0$) (which would indicate independence). With a \textit{p}-value of 0.001, we reject the null hypothesis, concluding that $C$ and $C_{\text{cand.}}$ exhibit significant dependence. (See Table \ref{tab: statistical validation} in the Appendix for full results.)

\begin{figure*}[h!]
    \centering
    \begin{subfigure}{0.32\textwidth}
        \centering
        \includegraphics[width=\linewidth]{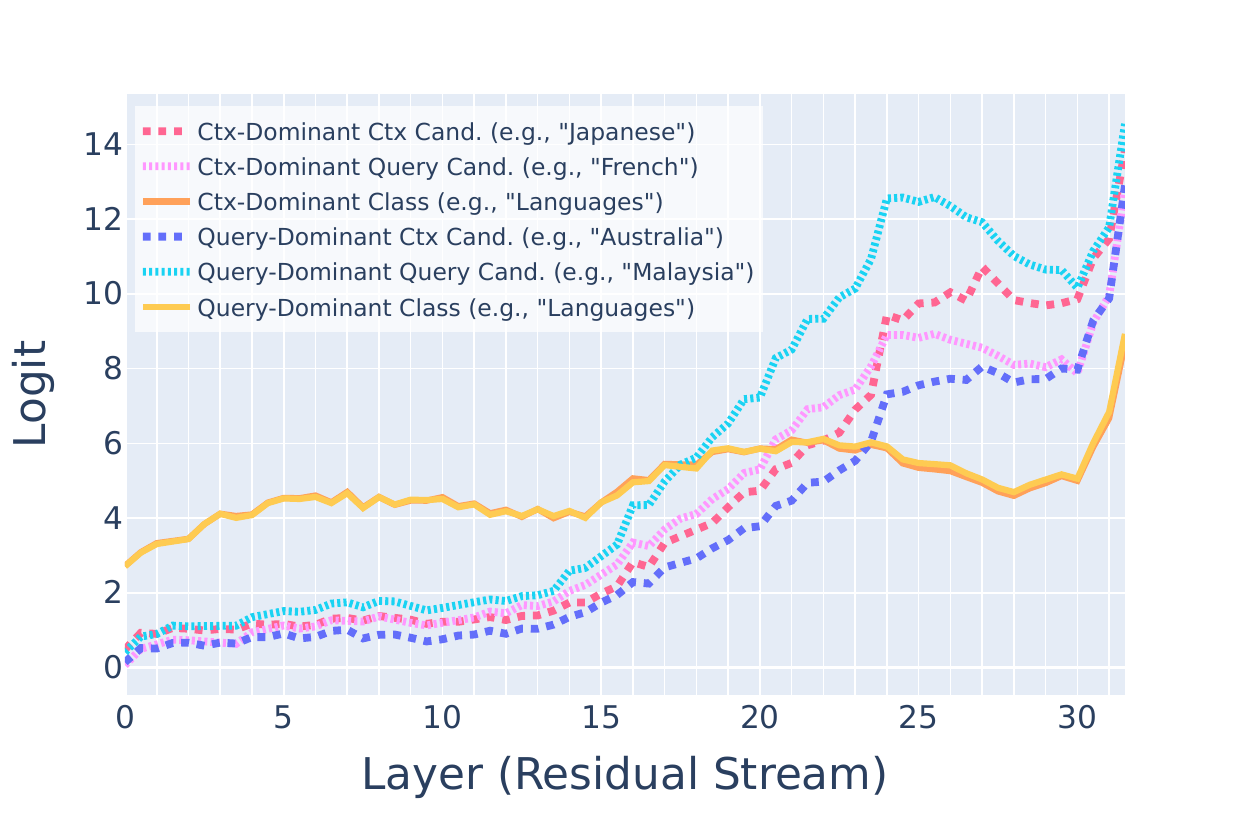}
        \caption{C+Q Token Logit (Llama)}
        \label{fig: c+q token logit}
    \end{subfigure}
    \begin{subfigure}{0.32\textwidth}
        \centering
        \includegraphics[width=\linewidth]{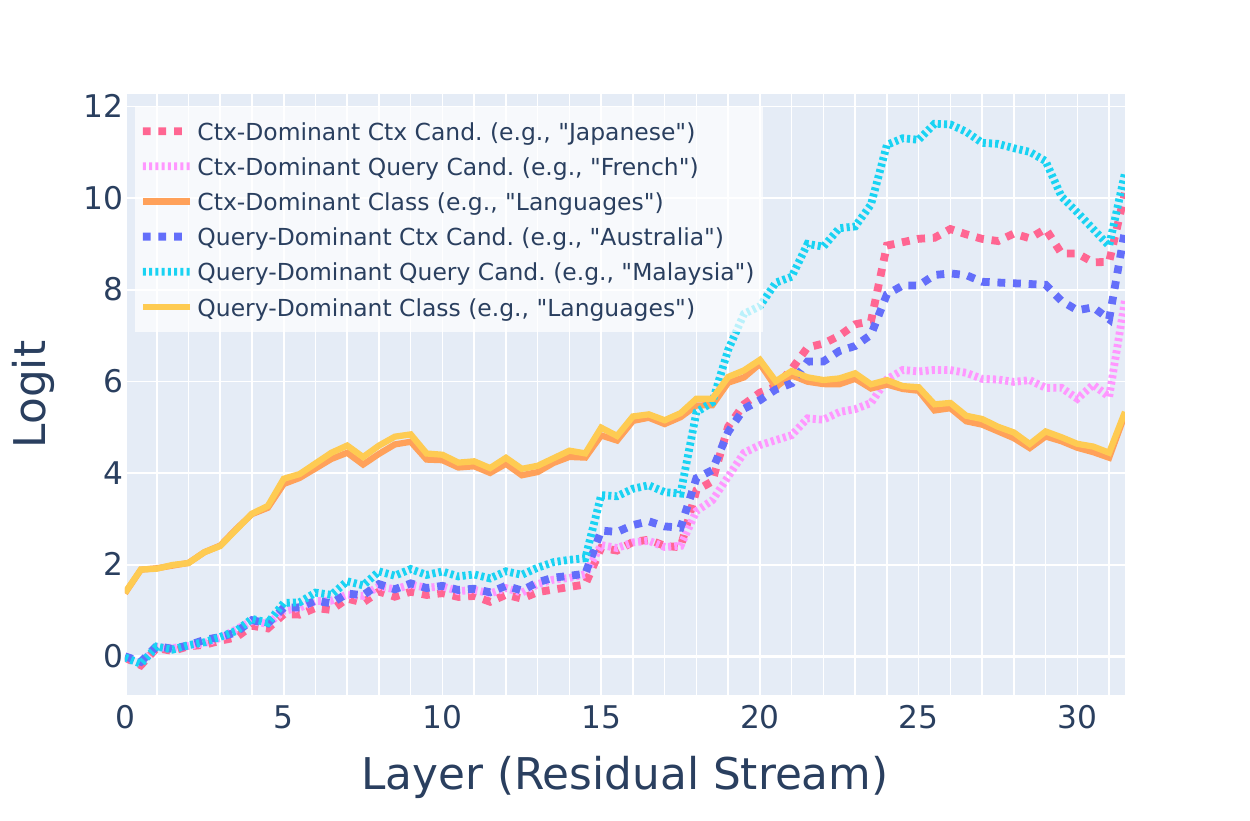}
        \caption{C+Q Token Logit (Mistral)}
        \label{fig:c+q token logit mistral}
    \end{subfigure}
    \begin{subfigure}{0.32\textwidth}
        \centering
        \includegraphics[width=\linewidth]{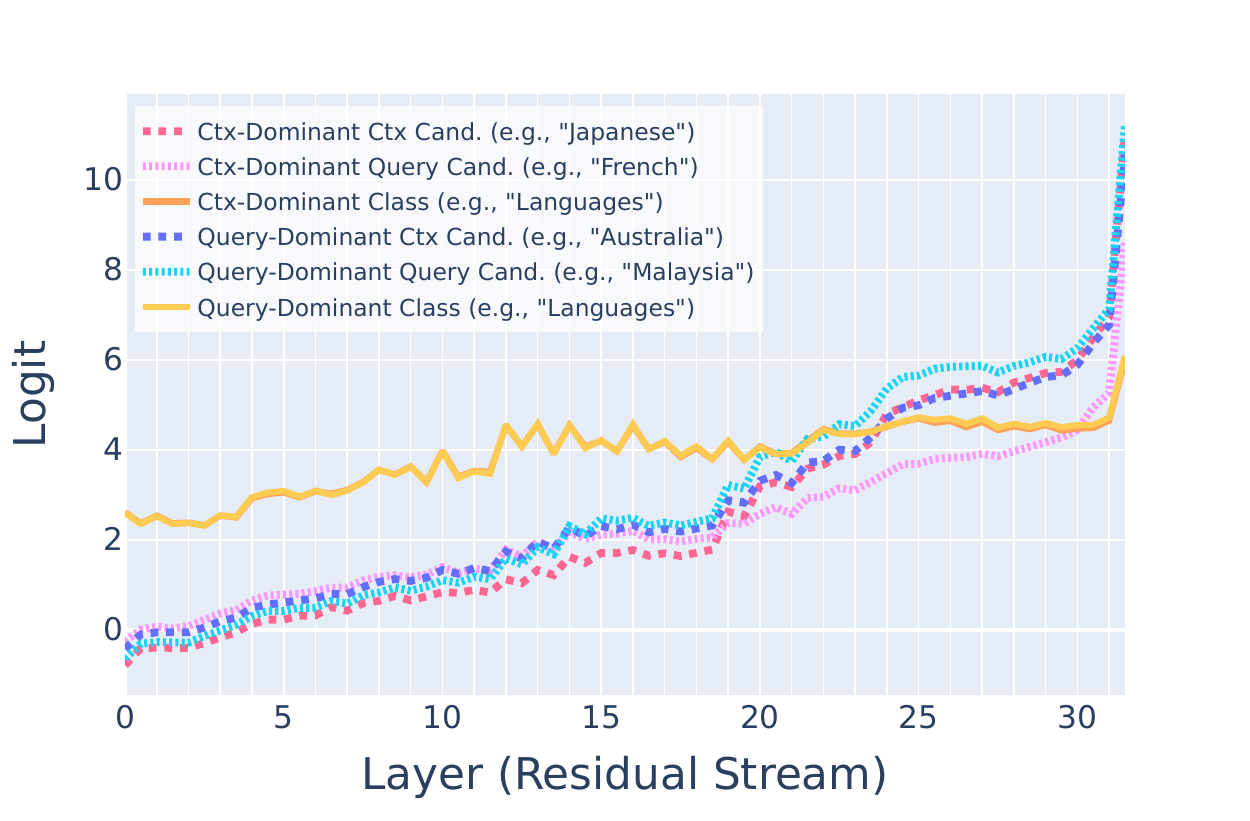}
        \caption{C+Q Token Logit (Pythia)}
        \label{fig:c+q token logit pythia}
    \end{subfigure}
    \caption{Logit attribution (C+Q condition) along residual stream ($R_{T,l}^{\text{1}}$, $R_{T,l}^{\text{2}}$) reveals the construction of {\color{YellowOrange}abstract class representation} in the lower layers, with competition between $Q_{\text{cand.}}$ (dashed) and $C_{\text{cand.}}$ (dotted) in the mid to higher layers. The example token in parenthesis correspond to Table~\ref{tab:dominant_examples}. Additional results are in Appendix \ref{appendix: logit-attribution}.}
    \label{fig: logit-attribution}
\end{figure*}

\begin{table*}[t]
    \centering
    \resizebox{2.1\columnwidth}{!}{%
    \begin{tabular}{@{}ccccccccccccccc@{}}
        \toprule
         \textbf{L16} & \textbf{L17} & \textbf{L18} & \textbf{L19} & \textbf{L20} & \textbf{L21} & \textbf{L22} & \textbf{L23} & \textbf{L24} & \textbf{L25} & \textbf{L27} & \textbf{L28} & \textbf{L29} & \textbf{L30} & \textbf{L31} \\
        \midrule
         \textbf{{\color{YellowOrange}languages}} & \textbf{{\color{YellowOrange}languages}} & /is & \textbf{{\color{YellowOrange}languages}} & \textbf{{\color{YellowOrange}languages}} & \textbf{{\color{YellowOrange}languages}} & \textbf{{\color{YellowOrange}languages}} & \textbf{{\color{YellowOrange}languages}} & English & English & English & English & English & English & English \\
         /is & /is & \textbf{{\color{YellowOrange}languages}} & /is & English & English & English & English & English & English & English & English & English & English & Japanese \\
        \bottomrule
    \end{tabular}
    }
    \caption{Logit lens on Llama-3 showing top-1 predictions shifting from abstract concepts (e.g., `languages') to concrete instances (e.g., `English' or `Japanese') across layers. The first and second row correspond to $R_{T,l}^{\text{1}}$, and the second row is $R_{T,l}^{\text{2}}$, respectively. See Appendix~\ref{appendix: logit_lens_example} for the corresponding prompt and associated probabilities.}
    \label{tab:logit_lens_example}
\end{table*}

\section{Mechanistic Analysis of Contextual Answers}
We next investigate whether the models' internal computations reflect the class-based generalization that we observed above. 
In Sec.~\ref{sec: logit_attribution}, we use logit attribution to show that models construct \textbf{abstract class representations}, supporting the class component of our hypothesis. In Sec.~\ref{sec: activation patching} and Sec.~\ref{sec: attention knockout}, we apply activation patching and attention knockout to reveal that feature selection in our hypothesis arise from \textbf{competition between circuits}, where distinct query-based pathways (computing $Q_{\text{cand.}}$) and context-based pathways (computing $C_{\text{cand.}}$) compete to determine the final answer. These findings provide mechanistic evidence for our hypothesis.

\paragraph{Data} We randomly draw 1,000 context-dominant and 1,000 query-dominant datapoints from case 2 and case 3 in Table \ref{tab:model_comparison} as these cases have both query- and context-based candidates.
\subsection{Logit Attribution}\label{sec: logit_attribution}
\paragraph{Method} To explore how models build answers across layers, we apply logit attribution \cite{nostalgebraist2020} to trace predictions across layers by projecting hidden states onto the vocabulary space. Given a prompt with $T$ tokens and a model with $L$ layers, we extract hidden states at the last token position $h_{T,j} \in \mathbf{R}^d$, where $j \in \{1,..., L\}$ and $d$ is the hidden size. These are projected onto the vocabulary space using $\text{Unembed}(\text{LayerNorm}(h_{T,j})) \in \mathbf{R}^{|V|}$, where the $\text{Unembed}$ matrix corresponds to the transpose of the input embedding weights.
Models maintain a residual stream for each token $i$, which accumulates information as it passes through each layer. At each layer, two key transformations occur: attention update ($A_{i,l}$) and MLP update ($M_{i,l}$). Mathematically, the updates follow:
\begin{align}
A_{i,l} &= \text{ATTN}(R_{i,l}^{\text{0}})\\
R_{i,l}^{\text{1}} &= A_{i,l} + R_{i,l}^{\text{0}} \label{eq: r1}\\
M_{i,l} &= \text{MLP}(R_{i,l}^{\text{1}})\\
R_{i,l}^{\text{2}} &= M_{i,l} + R_{i,l}^{\text{1}} \label{eq: r2}
\end{align}
where $R_{i,l}^{\text{1}}$ is the residual stream after attention at layer $l$,  and  $R_{i,l}^{\text{2}}$ is the final residual stream at layer $l$ after the MLP update. (See Appendix \ref{appendix: logit-attribution}.)

\paragraph{Findings} To understand how different tokens evolve across layers, we project the last token residual stream $R_{T,l}^{\text{1}}$ (after attention) and $R_{T,l}^{\text{2}}$ (after the MLP) onto the vocabulary space at each layer. Figure \ref{fig: logit-attribution} tracks the logits for $C_{\text{cand.}}$, $Q_{\text{cand.}}$, and class tokens under the $C+Q$ condition. Additional results are provided in Appendix \ref{appendix: logit-attribution}.
Figure \ref{fig: logit-attribution} reveals a hierarchical class-to-instance process in answer generation. Early layers prioritize class token logits (solid) like ``languages'', suggesting that the model first constructs abstract class representations. Around the middle layers, candidate answer logits (dashed/dotted) begin to rise, refining these abstract representations into concrete answers. In Table 
\ref{tab:logit_lens_example}, a concrete example of logit lens top-1 predictions reveals how Llama-3 shifts from abstract class to concrete instances. This pattern supports our hypothesis that \textbf{models leverage class-based information in shaping their predictions}.

Moreover, the figures highlight a \textbf{competition between $C_{\text{cand.}}$ (dashed) and $Q_{\text{cand.}}$ (dotted)}, particularly in context-dominant cases (pink). In early layers, logits for $C_{\text{cand.}}$ and $Q_{\text{cand.}}$ form two distinct groups, regardless of dominance. Around layer 14, $Q_{\text{cand.}}$ (dotted) in both cases begin to split, followed by $C_{\text{cand.}}$ (dashed) in layer 17. By layer 24, $C_{\text{cand.}}$ (dark pink) surpass $Q_{\text{cand.}}$ (light pink) logits in context-dominant settings, marking a decisive shift in the competition. After this, the early two-group pattern reemerges but with reversed dominance — context-based candidates prevail in context-dominant cases, and query-based candidates in query-dominant cases. By layer 29, the final prediction is fully formed, with the top logits corresponding to the final output. These observations reveal key insights: (i) \textbf{existence of competition}: even when the final prediction is query-dominant, context-based candidates remain actively computed across layers. (ii) \textbf{critical transition (Layers 17–24)}: the decisive competition between query- and context-based candidates occurs primarily in this range, determining which candidate is promoted.

\begin{figure*}[htbp]
    \centering
    \begin{subfigure}{0.48\linewidth}
        \centering
        \includegraphics[width=\linewidth]{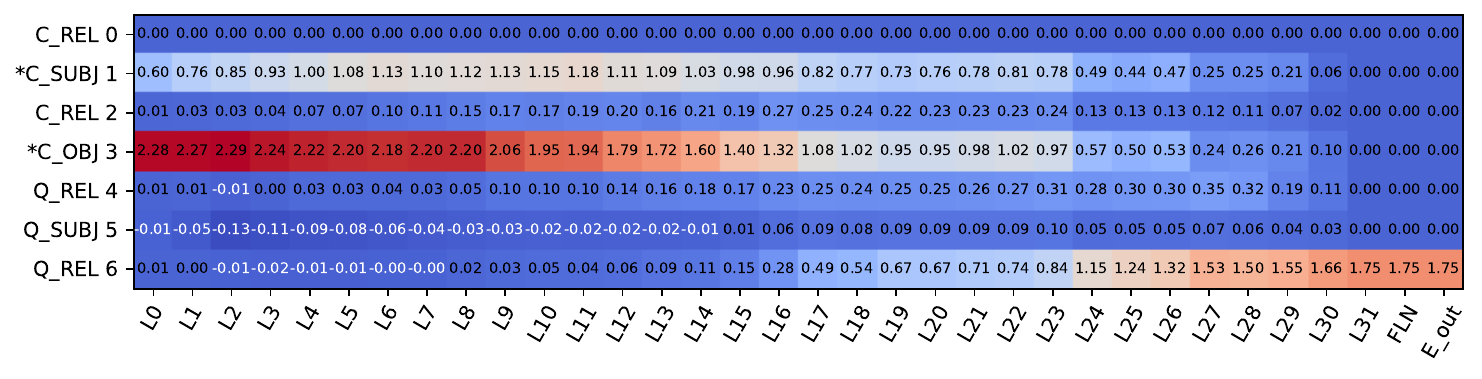}
        \caption{{\color{magenta}Context circuit} in context-dominant case.}
        \label{fig: ctx-driven-ctx-circuit-llama}
    \end{subfigure}
    \hfill
    \begin{subfigure}{0.48\linewidth}
        \centering
        \includegraphics[width=\linewidth]{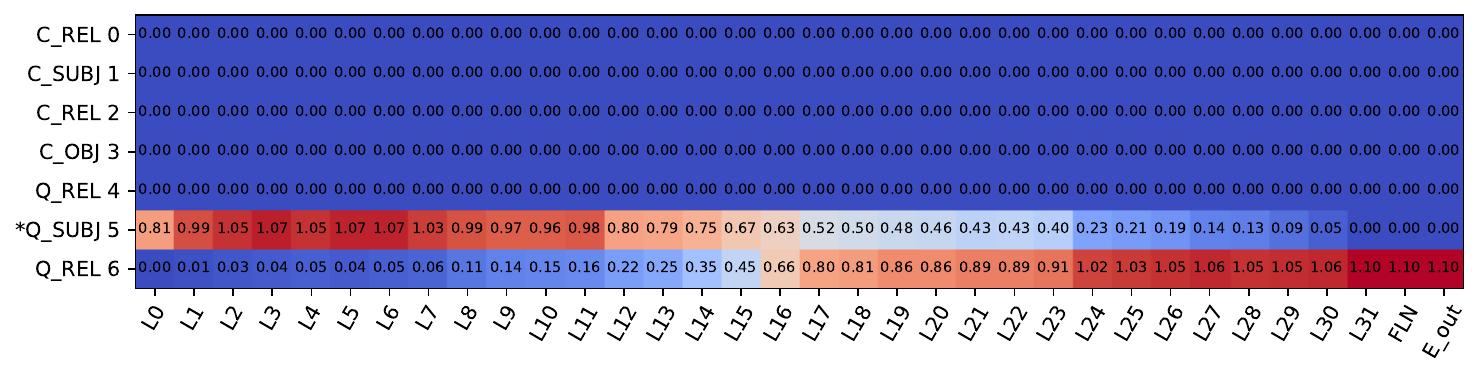}
        \caption{{\color{blue}Query circuit} in context-dominant case.}
        \label{fig: ctx-driven-query-circuit-llama}
    \end{subfigure}
    
    
    \begin{subfigure}{0.48\linewidth}
        \centering
        \includegraphics[width=\linewidth]{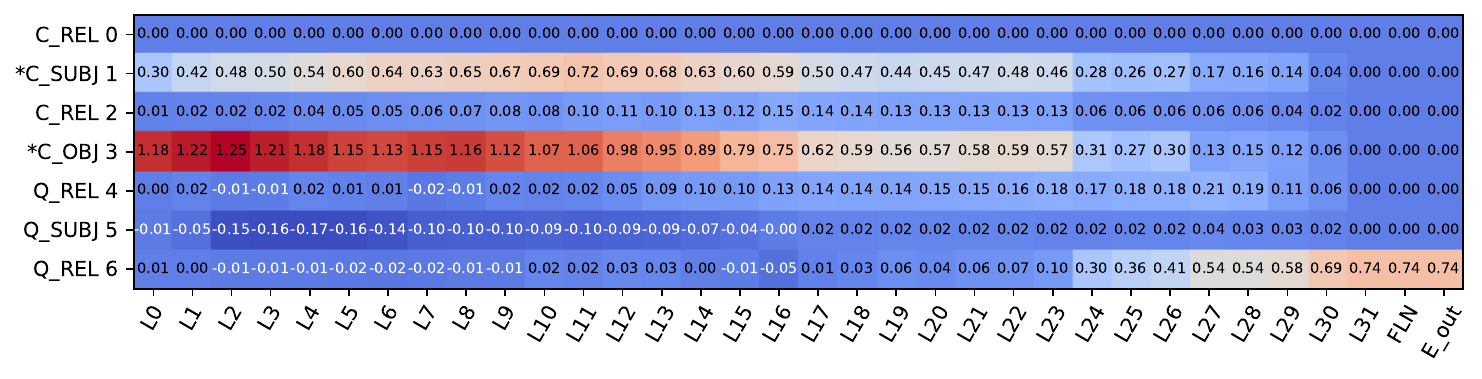}
        \caption{{\color{magenta}Context circuit} in query-dominant case.}
        \label{fig: query-dominant-ctx-circuit-llama}
    \end{subfigure}
    \hfill
    \begin{subfigure}{0.48\linewidth}
        \centering
        \includegraphics[width=\linewidth]{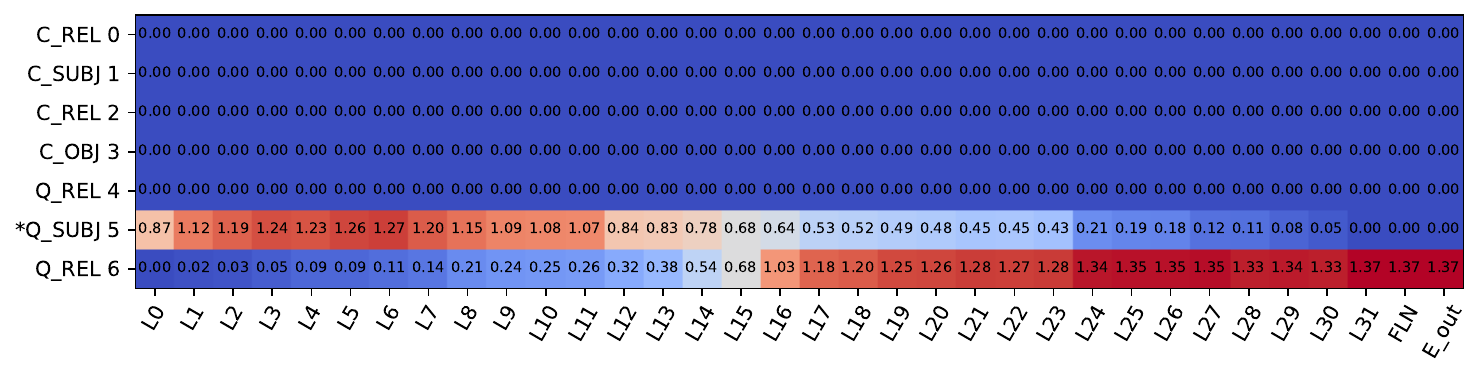}
        \caption{{\color{blue}Query circuit} in query-dominant case.}
        \label{fig: query-dominant-query-circuit-llama}
    \end{subfigure}
    \caption{Left-hand plots demonstrate the {\color{magenta}context circuit}, which extracts features from context and computes context-based candidates, while right-hand plots illustrate the {\color{blue}query circuit}. These circuits are the same in both context- and query-dominant cases; the difference lies in their strength, revealing the competition between context- and query-based candidates. An example is $\text{C}\_\text{REL0}=\text{[BOS]}$, {C\_SUBJ1}=` Honda Civic', {C\_REL2}=`, produced by',{C\_OBJ3}=` Honda', {Q\_REL4}=`. The original language of ',{Q\_SUBJ5}=` A Secret', {Q\_REL6}=` was'.}
    \label{fig:all-circuits}
\end{figure*}

\subsection{Activation Patching}\label{sec: activation patching}
\paragraph{Method} To understand the competition between $C_{\text{cand.}}$ and $Q_{\text{cand.}}$, we investigate whether distinct \textbf{context and query circuits} exist within the model's internal activations. We apply activation patching \cite{ghandeharioun2024patchscopes, meng2022locating}, a technique for causal intervention that selectively perturbs and restores activations to assess their contribution.
We conduct three model runs:
\textbf{(1) Clean run:} Standard forward pass with the original prompt, recording activations $\bigcup h_{i,l}^{0}$.
\textbf{(2) Corrupted run:} Forward pass with Gaussian noise $\epsilon \sim \mathcal{N}(0, \sigma$\footnote{We follow \citet{meng2022locating} to set $\sigma=0.3$ as three times of the empirical standard deviation of the input embeddings.})) injected into context or query topic token embeddings, yielding perturbed activations $\bigcup h_{i,l}^{1}$, and the final log-probabilities of candidates $\log p(t | \bigcup h_{i, l}^{1})$.
\textbf{(3) Restoration run:} Same as the corrupted run, but iterating over all $i$ and $l$, restoring each $h^{0}_{i^*, l^*}$, while keeping the rest corrupted. \\
By injecting noise at context subject and object (\textbf{context patching}) \textit{or} query subject position (\textbf{query patching}) and measuring the recovery of predictions, we differentiate \textbf{context and query circuits}, tracing how features from these tokens propagate through the model and how they contribute to context-based or query-based candidates.

The \textbf{restoration effect} for each $i^*$ and $l^*$ is calculated as in Eq.~\ref{RE-equation},  where $t$ $\in$ $\{\text{$C_{\text{cand.}}$, $Q_{\text{cand.}}$} \}$, with higher values indicate stronger contributions. \begin{align}
\text{RE}(i^*, l^*, t) &=  \log p(t | h_{i^*, l^*}^{0} \cup h_{-i^*, -l^*}^{1}) \nonumber \\
&\quad - \log p(t | \bigcup h_{i, l}^{1})  
\label{RE-equation} \\
\Delta_{\text{query}}(i^*,l^*) &= \text{RE}(i^*,l^*, Q_{\text{cand.}}) - \nonumber \\
&\quad \text{RE}(i^*,l^*, C_{\text{cand.}}) 
\label{query-patching} \\
\Delta_{\text{context}}(i^*,l^*) &= \text{RE}(i^*,l^*, C_{\text{cand.}}) - \nonumber \\
&\quad \text{RE}(i^*,l^*, Q_{\text{cand.}})  
\label{ctx-patching}
\end{align}

In context and query circuits, we compute Eq.~\ref{ctx-patching} (Figures ~\ref{fig: ctx-driven-ctx-circuit-llama}, \ref{fig: query-dominant-ctx-circuit-llama}) and Eq.~\ref{query-patching} (Figures~\ref{fig: ctx-driven-query-circuit-llama}, \ref{fig: query-dominant-query-circuit-llama}), respectively. Comparing restoration effects maps circuits responsible for context- and query-based candidates and identifies where their competition occurs. (See Appendix \ref{appendix: activation-patching} for implementation details and additional results.)

\begin{table}[h]
    \centering
    \small
    \setlength{\tabcolsep}{4pt} 
    \renewcommand{\arraystretch}{0.8} 
    \begin{tabular}{lcclcc}
        \toprule
        & Orig. & \multicolumn{2}{c}{L17+L24} & \multicolumn{2}{c}{2 Rand.} \\
        \cmidrule(lr){2-2} \cmidrule(lr){3-4} \cmidrule(lr){5-6}
        & Prob. & Prob. & $\Delta$ & Prob. & $\Delta$ \\
        \midrule
        \multicolumn{6}{c}{\textbf{Context-Dominant}} \\
        \midrule
        $C_{\text{cand.}}$   & 25.5 & \textbf{13.1} & -12.4 & 21.0 & -4.5 \\
        $Q_{\text{cand.}}$ & 8.6  & \textbf{14.8} & + 6.2  & 10.8 & +2.2 \\
        \midrule
        \multicolumn{6}{c}{\textbf{Query-Dominant}} \\
        \midrule
        $Q_{\text{cand.}}$ & 35.2 & \textbf{26.8} & -8.4  & 29.6 & -5.7 \\
        $C_{\text{cand.}}$   & 6.6  & \textbf{11.3} & +4.7  & 7.4  & +0.8 \\
        \bottomrule
    \end{tabular}
    \caption{Effect of attention knockout on context- ($C_{\text{cand.}}$) and query-based ($Q_{\text{cand.}}$) candidate probabilities on Llama-3. ``Orig.'' = No intervention, ``2 Rand.'' = Average of interventions on two random layers over three runs.  $\Delta$ denotes the change from the original setting.}
    \label{tab:knockout_results_llama}
\end{table}

\paragraph{Findings} The results reveal distinct circuits for context- and query-based pathways. Figures \ref{fig: ctx-driven-ctx-circuit-llama} and \ref{fig: query-dominant-ctx-circuit-llama} show the same context circuit aggregating information from the context subject and object in both cases, transferring it to the final token position from layer 17 onward. In contrast, Figures \ref{fig: ctx-driven-query-circuit-llama} and \ref{fig: query-dominant-query-circuit-llama} indicate that the same query circuit for both cases integrating query subject information earlier than in the context circuit, from layer 8. The log-probability increases after layer 24 in context circuit and after layer 16 in query circuit.

Both circuits exist across context- and query-dominant cases, but their relative strength determines the final prediction. In context-dominant cases, the context circuit wins, with a larger log-probability difference (max 2.28) compared to the query circuit (max 1.10). Conversely, in query-dominant cases, the query circuit exerts a stronger influence (max 1.37 vs. 1.25). Notably, between layers 17 and 24, the query-dominant case shows minimal context information transfer (Figure \ref{fig: query-dominant-ctx-circuit-llama}), aligning with slower logit attribution growth (Figure \ref{fig: logit-attribution}). This confirms that both pathways exist for both cases with final predictions depending on their relative activation strength, and layers 17 to 24 are the key to promoting context-based candidates.

\subsection{Flipping Model Predictions via Attention Knockout}\label{sec: attention knockout}
To examine the causal role of internal competition in shaping the final output $A_{C+Q}$, we intervene in two key layers of the context circuit: layer 17 (where context first transfers to the last token) and layer 24 (where it is most integrated). By restricting attention to the query in the context-dominant case and to the context in the query-dominant case, we test whether predictions can be flipped (e.g., ``Japanese'' to ``French''). See Appendix~\ref{appendix: attention knockout} for details and additional results. Table \ref{tab:knockout_results_llama} (Llama) shows that in the context-dominant case, blocking context flow causes $Q_\text{cand.}$ probabilities to surpass $C_\text{cand.}$ on average, flipping 465/1000 datapoints to query-based candidates. In the query-dominant case, intervention increases $C_\text{cand.}$ probability by 4.7 and decreases $Q_\text{cand.}$ probability by 8.4, flipping 225/1000 datapoints. These results confirm the competition between  $C_\text{cand.}$ and $Q_\text{cand.}$, and that these two layers are the key to promoting context-based candidates.

\begin{table*}[h!]
\centering
\small
\begin{tabular}{@{}p{2.5cm}p{13.2cm}@{}}
\toprule
\textbf{Model Family} & \textbf{Context Demonstration + Query and Answer} \\
\midrule

\multirow{6}{*}{Pythia 12B} 

& 
\textbf{Prompt:} {\color{blue}Davie Fulton} found employment in {\color{blue}Ottawa}. \textit{Valiant Lady} premiered on \_\_\_\_\_ \newline
\textbf{Answer:} $A_{C+Q}=$ {\color{blue}CBC}, $A_Q=$ September \\ 

\cmidrule(l){2-2}

& 
\textbf{Prompt:} {\color{blue}{1 fille \& 4 types}} was written in {\color{blue}French}. Cool \& Dre, founded in \_\_\_\_\_ \newline
\textbf{Answer:} $A_{C+Q}=$ {\color{blue}Paris}, $A_Q=$ Compton \\

\cmidrule(l){2-2}

& 
\textbf{Prompt:} {\color{blue}Cologne Cathedral}, which is named after Peter. Montana borders with \_\_\_\_\_ \newline
\textbf{Answer:} $A_{C+Q}=$ {\color{blue}Germany}, $A_Q=$ Wyoming \\

\midrule

\multirow{6}{*}{LLama-3 70B} 

& 
\textbf{Prompt:} {\color{blue}{Ilm al-Kalam}} is a part of {\color{blue}Islam}. \textit{The Man-Machine} was written in \_\_\_\_\_ \newline
\textbf{Answer:} $A_{C+Q}=$ {\color{blue}Arabic}, $A_Q=$ English \\ 

\cmidrule(l){2-2}

& 
\textbf{Prompt:} {\color{blue}Diarmuid Martin}, who holds the position of bishop. Riga is a twin city of \_\_\_\_\_ \newline
\textbf{Answer:} $A_{C+Q}=$ {\color{blue}Dublin}, $A_Q=$ Cal \\ 

\cmidrule(l){2-2}

& 
\textbf{Prompt:} {\color{blue}David Hatch}, who is employed by {\color{blue}BBC}. Rudolf Lothar passed away at \_\_\_\_\_ \newline
\textbf{Answer:} $A_{C+Q}=$ {\color{blue}London}, $A_Q=$ New \\

\bottomrule
\end{tabular}
\caption{Examples of context-based candidates on larger model sizes (Pythia 12B and Llama-3 70B).}
\label{tab:qual-examples-larger-models}
\end{table*}

\paragraph{Summary} These findings support the class-based (mis)generalization hypothesis. Logit attribution confirms that models first construct abstract class representations before refining them into specific answers. Activation patching reveals competing circuits for feature selection: one favoring direct query-based pathway and the other integrating contextual cues, with their strength shaping the final output. Notably, context circuit strengthens between layers 17 and 24, validated by the flipped predictions from attention knockout.

\section{Discussion}
\subsection{Does scale alone alleviate irrelevant context hallucinations?}
To test whether scaling up model size naturally mitigates irrelevant context hallucinations — and thereby reduces class-based generalization — we evaluate the largest models available within our resource budget: Pythia 12B and LLaMA-3 70B. Experimental details are provided in Appendix~\ref{app: larger-models-experiment-details}. Contrary to the hypothesis that scale might resolve this issue, results in Table~\ref{tab:top3-composition-larger-models} show that class-based generalization persists with similar frequency as in 7B/8B models (Sec.~\ref{sec: behaviour_changes}). Table~\ref{tab:qual-examples-larger-models} provides qualitative examples. The statistical test (Appendix Table~\ref{tab:statistical-pmi})—analogous to Sec.~\ref{sec: testing}—confirms that correlations between context and context-based candidates remain significant even at larger scales.

\subsection{Is the class-based generalization phenomenon sensitive to the prompt templates?}
To test for prompt sensitivity, we conduct the Llama-3 8B experiments using alternate templates sampled from the ParaRel dataset (See Appendix Table~\ref{tab: pararel_dataset_template_variation}). As shown in Table~\ref{tab:top3-composition-larger-models}, the phenomenon persists with similar frequencies, indicating that prompt wording has no major impact. Statistical validation (Appendix Table~\ref{tab:statistical-pmi}) again confirms statistically significant context influence, consistent with our earlier findings (Sec.~\ref{sec: testing}).

\begin{table}[h]
    \centering
    \small
    \begin{tabular}{@{}p{1cm}p{2.5cm}p{0.8cm}p{0.8cm}p{1cm}@{}}
        \toprule
        \textbf{Case} & \textbf{Top-3 Candidates} & \textbf{Llama-70B} & \textbf{Pythia-12B} & \textbf{Llama-8B-Prompt} \\
        \midrule
        No influence & 1. All query-based ($C_{\text{cand.}}= \emptyset$)  & 49.0\% & 39.6\% & 48.7\%\\
        \midrule
        \multirow{2}{=}{Query-dominant}  & 2. Mix, top-1 is query-based & 27.2\%  & 28.0\% & 27.7\%\\
        \midrule
        \multirow{4}{=}{Context-dominant} 
        & 3. Mix, top-1 is context-based & 14.5\%  & 19.0\% & 13.8\% \\
        \cmidrule{2-5}
        & 4. All context-based ($Q_{\text{cand.}}= \emptyset$) & 9.3\% & 13.4\% & 9.8\% \\
        \bottomrule
    \end{tabular}
    \caption{Breakdown of samples according to the composition of $A_{C+Q}^{\text{top-3}}$ for Llama-70B, Pythia 12B and Llama-8B with prompt template variations.}
    \label{tab:top3-composition-larger-models}
\end{table}

\section{Conclusion}
By analyzing the mechanism behind irrelevant context hallucinations, our study demonstrates that LLMs exhibit class-based (mis)generalization, relying on abstract class structures in a systematic yet flawed manner. Through mechanistic analysis, we show that this phenomenon arises from hierarchical class-to-instance predictions and competing circuits that mediate feature selection. These findings challenge a potential misconstrual of the stochastic parrot hypothesis that LLMs can only regurgitate surface-level patterns. Rather, we argue that LLMs are \emph{``stochastic chameleons''} -- they exhibit generalization by leveraging class structures and dynamically adapting their responses to contextual cues, in ways that are neither purely memorized nor necessarily reliable.

\section{Limitations}
Our work has several limitations. First, our experiments are conducted in a controlled setting, which helps isolate generalization from memorization and enables analysis at both behavioral and mechanistic levels. However, future work could improve upon this by designing setups that disentangle memorization and generalization in naturally occurring text. Second, our study is limited to English-language datasets, and we only evaluate models of certain sizes (around 7–8B, 12B and 70B). Do smaller models also display class-based generalization, and if so, what is the minimum size required? Third, in the mechanistic interpretability section, we focus primarily on layer-wise analysis to support our main hypothesis, while attention head analysis is left for future work. Finally, while we conduct interventions, our primary goal is not to mitigate contextual hallucinations. Developing mitigation methods informed by our findings and evaluating their effectiveness is an important direction for future research.

\section*{Acknowledgments} We thank Cesare Spinoso-Di Piano and Zichao Li for helpful discussions. We are also grateful to the reviewers for their valuable comments and suggestions. Ziling Cheng is supported by a Fonds de Recherche
du Québec Nature et Technologies master research
scholarship (File \# 365574). Jackie Chi Kit Cheung is supported by the Canada CIFAR AI Chair program. We acknowledge the material support of NVIDIA for providing computational resources.

\bibliography{main}

\clearpage
\newpage
\appendix

\section{Dataset}\label{appendix: dataset}
The detailed breakdown of ParaRel dataset \cite{elazar-etal-2021-measuring} based on relation type is presented in Table \ref{tab: pararel_dataset}. We categorize the sub-datasets into 5 knowledge types based on the expected class or type of the answer (column `Ctx Type'): `language', `place', `company', `job', and if a sub-dataset doesn't fit into the above types then it is categorized as `others'. This is the dataset that we use for $Q$-only experiments, and we construct the dataset for $C+Q$ experiments by generating 3900 context variations spanning all knowledge types per query, resulting in a dataset of 106.2M data points. For each generation, we restrict the vocabulary to the set of tokens that begins with a capitalized English letter \cite{yu-etal-2024-mechanistic}. When evaluating, we lowercase generated and gold answers and perform string matching: if the top-1 generated answer is a substring of the gold answer, then this is correct. For evaluation, we set the temperature to zero for
all models to reduce output randomness.

\section{Class-based Generalization}
\label{appendix: hypothesis}
We further categorize class-based generalization into two distinct cases:
\begin{itemize}
    \item \textbf{Copying:} When a token belonging to the expected class appears in the context, the model is more likely to directly copy it as the answer. From a dataset statistics perspective, we observe a high copy rate when the context contains tokens belonging to the same class as the query. 
    
    \textbf{Example:} The mother tongue of Dominique Sanda is {\color{blue}French}. The original language of Puss in Boots was → {\color{blue}French}.
    \item \textbf{Non-copying:} When tokens of the expected query class are not explicitly present in the input, the model combines the expected class with relevant features inferred from context or query to generate an answer.
    
    \textbf{Example:} Honda Civic (fifth generation), produced by {\color{blue}Honda}. The original language of Tow Truck Pluck was → {\color{blue}Japanese}.
\end{itemize}


\section{Behavioral Changes Induced by Irrelevant Context}
\subsection{Irrelevant Context Hallucination Evaluation}
In Table \ref{tab:contextual_hallucinations_stats_other_model}, we provide detailed statistics of the accuracy/wrong rate for each model under each case for all three models. 

\begin{table}[h]
\centering
\small
\begin{tabular}{l l l l l l}
\toprule
\multirow{2}{*}{\textsc{Model}} & \multicolumn{2}{c}{\textsc{Q-only}} & \multicolumn{3}{c}{\textsc{C+Q}} \\
\cmidrule(lr){2-3} 
\cmidrule(lr){4-6}
& \textsc{Case} & \textsc{Prop.} &  \textsc{Case} & \textsc{Prop.} & \textsc{$\Delta$ Rate} \\
\cmidrule(lr){1-6}
\multirow{4}{*}{Llama}  
& \multirow{2}{*}{T}  & \multirow{2}{*}{47.2\%} & T $\rightarrow$ T & 35.7\% & 0\% \\
&                       &                       & T $\rightarrow$ F  &  11.5\% & 100\% \\
\cmidrule(lr){2-6}                     
& \multirow{2}{*}{F}  & \multirow{2}{*}{52.8\%} & F $\rightarrow$ T &  7.4\% & 100\% \\
&      &         & F $\rightarrow$ F & 45.4\% & 42.7\%\\
\cmidrule(lr){2-6}
 & Total  & \textbf{47.2\%}  &  Total & \textbf{43.1\%} & \textbf{38.3\%} \\
\cmidrule(lr){1-6}
\multirow{4}{*}{Mistral}  
& \multirow{2}{*}{T}  & \multirow{2}{*}{38.2\%} & T $\rightarrow$ T & 29.4\% & 0\% \\
&                       &                       & T $\rightarrow$ F  &  8.8\% & 100\% \\
\cmidrule(lr){2-6}                     
& \multirow{2}{*}{F}  & \multirow{2}{*}{61.8\%} & F $\rightarrow$ T &  5.9\% & 100\% \\
&      &         & F $\rightarrow$ F & 55.9\% & 59.5\%\\
\cmidrule(lr){2-6}
 & Total  & \textbf{38.2\%}  &  Total & \textbf{35.3\%} & \textbf{48.0\%} \\
\cmidrule(lr){1-6}

\multirow{4}{*}{Pythia}  
& \multirow{2}{*}{T}  & \multirow{2}{*}{30.9\%} & T $\rightarrow$ T & 22.4\% & 0\% \\
&                       &                       & T $\rightarrow$ F  &  8.4\% & 100\% \\
\cmidrule(lr){2-6}                     
& \multirow{2}{*}{F}  & \multirow{2}{*}{69.1\%} & F $\rightarrow$ T &  5.6\% & 100\% \\
&      &         & F $\rightarrow$ F & 63.6\% & 67.8\%\\
\cmidrule(lr){2-6}
 & Total  & \textbf{30.9\%}  &  Total & \textbf{28.0\%} & \textbf{57.1\%} \\

\bottomrule
\end{tabular}
\caption{Comparison of proportions (Prop.) of correct and incorrect answers in Q-only and C+Q cases, along with answer change rates ($\Delta$ Rate) for different models. Average across 39 datasets are reported. In the `Total' row, under `Prop.' column, it indicates the global accuracy across different cases, while under `$\Delta$ Rate' column, it underlies the global answer change rate.}
\label{tab:contextual_hallucinations_stats_other_model}
\end{table}

Table \ref{tab:contextual_hallucinations_stats_other_model} shows that models are not robust against irrelevant context. Even when a single irrelevant demonstration is prepended, models exhibit notable shifts in performance. For instance, in Llama, 11.5\% of previously correct answers become incorrect, while 7.4\% of incorrect answers are corrected after adding context. However, accuracy alone does not capture all behavioral shifts — predictions can still change even if they remain incorrect. 


\subsection{Composition of $A_{C+Q}^{\text{top-3}}$}
Table \ref{tab:model_comparison_detailed} provides counts and proportions of the breakdown of samples according to the composition of $A_{C+Q}^{\text{top-3}}$ for three models
based on 106M datapoints.

\begin{table*}[h]
    \centering
    \small
    \begin{tabular}{@{}p{1cm}p{6cm}p{2cm}p{2cm}p{2cm}@{}}
        \toprule
        \textbf{Case} & \textbf{Top-3 Candidates} & \textbf{Llama} & \textbf{Mistral} & \textbf{Pythia}\\
        \midrule
        \multirow{2}{=}{No influence} 
        & 1. All query-based  & 50,874,341 (47.9
        \%)& 51,013,564 (48.0\%) & 41,833,760 (39.3\%)\\
        \cmidrule{1-5}
        \multirow{2}{=}{Query-dominant} & 2. Mix: Query + Context, top-1 is query-based & 27,940,495 (27.9
        \%) & 27,342,287 (25.7\%) & 28,885,252 (27.2\%)\\
        \midrule
        \multirow{4}{=}{Context-dominant} 
        & 3. Mix: Query + Context, top-1 is context-based & 16,069,253 (15.1
        \%) & 17,013,397 (16.0\%) & 20,412,292 (19.2\%)\\
        \cmidrule{2-5}
        & 4. All context-based & 11,353,892 (10.1
        \%) & 10,963,675 (10.3\%)  & 15,250,026 (14.3\%)\\
        \bottomrule
    \end{tabular}
    \caption{Breakdown of samples according to the composition of $A_{C+Q}^{\text{top-3}}$, based on 106M datapoints.} 
    \label{tab:model_comparison_detailed}
\end{table*}

\section{Annotation}\label{appendix: annotation}

\subsection{Annotation Procedure}
To systematically evaluate the impact of irrelevant context on model predictions, we perform an annotation procedure for context-based candidates — those predictions that were influenced by the inclusion of extraneous context. The aim was to rigorously assess whether (i) these predictions incorporated identifiable features from the context, \textbf{and} (ii) appropriately combined them with the expected class as indicated by the query. 

\textbf{Step 1: Candidate Selection}
We first randomly sample a set of 500 context-based candidates from different sub-datasets, ensuring a diverse set of instances. Context-based candidates were selected for both context- and query-dominant cases.

\textbf{Step 2: Context Feature Identification}
For each context-based candidate, we analyzed the context —specifically the subject and object — to identify any features that could have been leveraged by the model in generating the response. (`context-influenced?' row in Table~\ref{tab:annotation_examples}).

Each feature is categorized as identifiable if it can be explicitly extracted from the context. For example, the country of origin of a figure (e.g., candidates `South' `Korea' for context subject `Lee Jong-hyun' in Example 5 in  Table~\ref{tab:annotation_examples}), country/continent of a district (`India', `Asia' for context object `Bihar' in Example 4 in Table~\ref{tab:annotation_examples}) are classified as \textit{identifiable}. In contrast, context-based candidates `Bee', `Beach' are categorized as non-context influenced for context subject `Grant Green' and object `jazz' as shown in Example 6 in Table~\ref{tab:annotation_examples}.

We ensure transparency by documenting the rationale. For example, in Example 2 of Table~\ref{tab:annotation_examples}, we provide the justification that `Svend Asmussen' is a Danish violinist and jazz musician, which supports that `Danmark' is a context-influenced candidate.

\textbf{Step 3: Class Verification}
Next, each context-based candidate is classified according to the abstract class suggested by the query. The candidate is compared to the expected class, and we verify whether the response falls within the correct category. For example, the context-based candidates `Vietnamese' and `Thai' for Example 1 in Table~\ref{tab:annotation_examples} have the correct class `language', but `South', `Korea' in Example 5 in Table~\ref{tab:annotation_examples} do not have the correct class because the query is asking about continent, not country.

\textbf{Step 4: Hypothesis Verification}
Finally, a context-based candidate is considered to satisfy the hypothesis if it meets the criteria from both Step 2 (context feature identification) and Step 3 (class verification). Only candidates that successfully integrate context features \textbf{and} align with the expected class are retained as valid instances.



\subsection{Annotation Examples}
Details of examples and non-examples are shown in Table \ref{tab:annotation_examples}.

\section{Statistical Validation of Contextual Influence}\label{appendix: testing}

Mean PMI values for each model are presented in Table \ref{tab: statistical validation}. A mean PMI of around 70 across all models and expected classes confirms strong statistical dependence (Table \ref{tab: statistical validation}).


Full results on three models in Table \ref{tab: statistical validation}.


\begin{table}[h!]
\centering
\small
\begin{tabular}{lccc}
\toprule
Value & \textbf{Llama-3} & \textbf{Mistral} & \textbf{Pythia} \\
\midrule
Mean PMI & 3.9 & 3.7 & 3.8  \\
T-statistic & 8.1 & 7.3 & 6.6\\
\textit{p}-value & 0.0006 & 0.0009 & 0.001\\
\bottomrule
\end{tabular}
\caption{Mean PMI values and T-test results for all three models.}
\label{tab: statistical validation}
\end{table}

\section{Logit Attribution}\label{appendix: logit-attribution}
\subsection{Implementation Details} When the target candidates or class have multiple tokens, we take the maximum logit, and average this maximum logit across all data points in the dataset. 

To obtain the class logits from the model, we predefine a list of tokens according to the relation type. 
\begin{itemize}
    \item Languages: languages, language, tongue, tongues, lingua, dialect, dialects
    \item Places: country, countries, place, places, location, locations, territory, city, cities, town, towns, village, villages, state, states, province, provinces, district, districts, continent, continents
    \item Companies: company, companies, manufacturer, manufacturers, make, firm, firms, business, corporation, corporations, enterprise, enterprises, organization, organizations, channel, channels, broadcaster, broadcasters, industry, industries
    \item Jobs: position, positions, job, jobs, career, careers, profession, professions, occupation, occupations, role, roles, assignment, assignments, employment, employments
    \item Others: expertise, area, areas, field, fields, subject, subjects, instrument, instruments, genre, music, religion, religions, concept, concepts, framework, frameworks, artifact, artifacts, type, types, part, parts, class, classes, eponym, eponyms, entity, entities, person, persons, place, places
\end{itemize}

\subsection{Logit Lens Example}\label{appendix: logit_lens_example}
\begin{figure*}
        \centering
        \includegraphics[width=\linewidth]{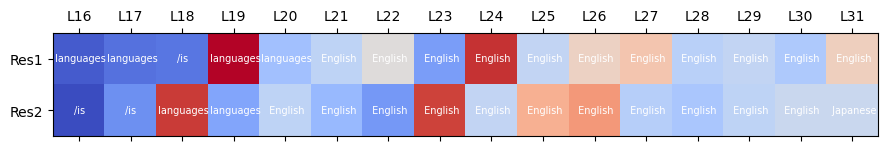}
        \caption{Logit lens on Llama-3 shows how model's top-1 predictions shift along the residual stream from abstract concepts (e.g., 'languages') to concrete instances (e.g., 'English' or 'Japanese') across layers. Red indicates high probability.}
        \label{fig: logit lens demo}
\end{figure*}

We provide an example of how the model's top-1 predictions shift along the residual stream from abstract concepts to concrete instances across layers in Table~\ref{tab:logit_lens_example} and Figure~\ref{fig: logit lens demo}. The prompt used is \textit{Honda Civic (fifth generation), produced by Honda. The original language of Tow Truck Pluck was}. Red indicates probability around 80\%. We show predictions above layer 15 because lower than this, the predictions are not interpretable.

\subsection{Additional Logit Attribution Results}\label{appendix: logit-attribution-llama}
\begin{figure*}[h!]
    \centering
    \begin{subfigure}{0.48\textwidth}
        \centering
        \includegraphics[width=\linewidth]{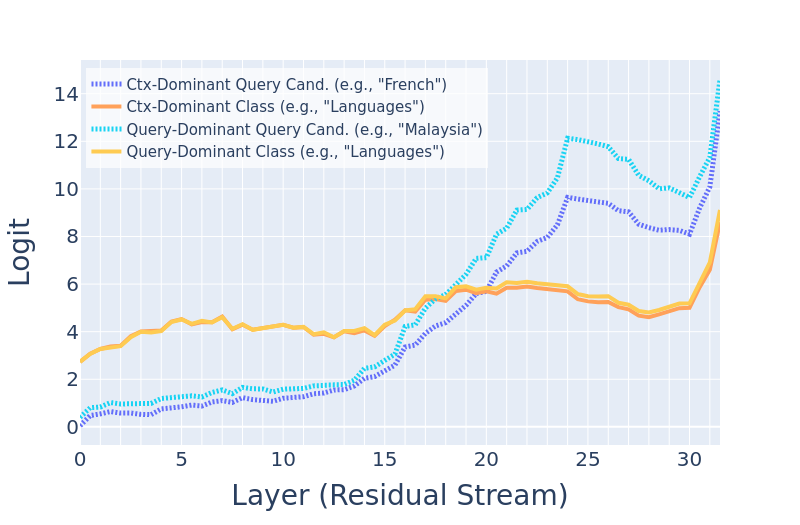}
        \caption{\textbf{Q-only:} Token logit from accumulated residual stream. ($R_{T,l}^{\text{1}}$, $R_{T,l}^{\text{2}}$) are visualized per layer.}
        \label{fig:q-only token logit llama}
    \end{subfigure}
    \begin{subfigure}{0.48\textwidth}
        \centering
        \includegraphics[width=\linewidth]{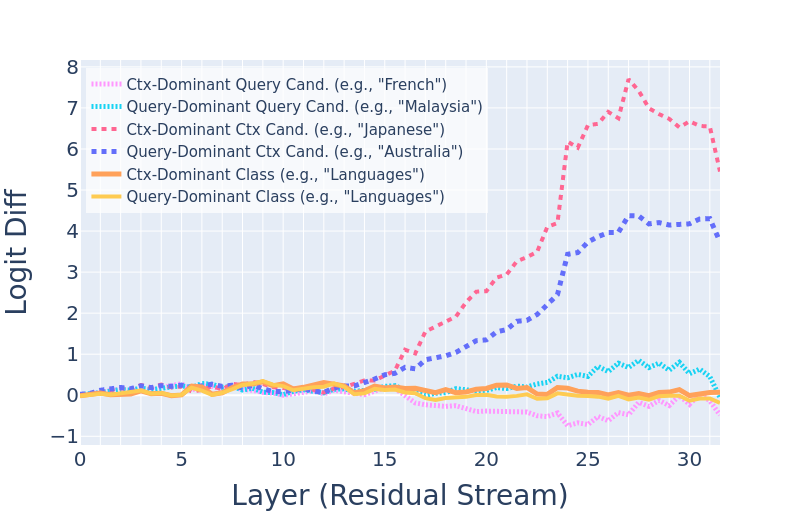}
        \caption{Token logit difference (Logit in \textbf{C+Q} - Logit in \textbf{Q-only}) from accumulated residual stream ($R_{T,l}^{\text{1}}$, $R_{T,l}^{\text{2}}$).}         
        \label{fig:c+q-q logit diff llama}
    \end{subfigure}
    \begin{subfigure}[b]{0.48\textwidth} 
        \centering
        \includegraphics[width=\textwidth]{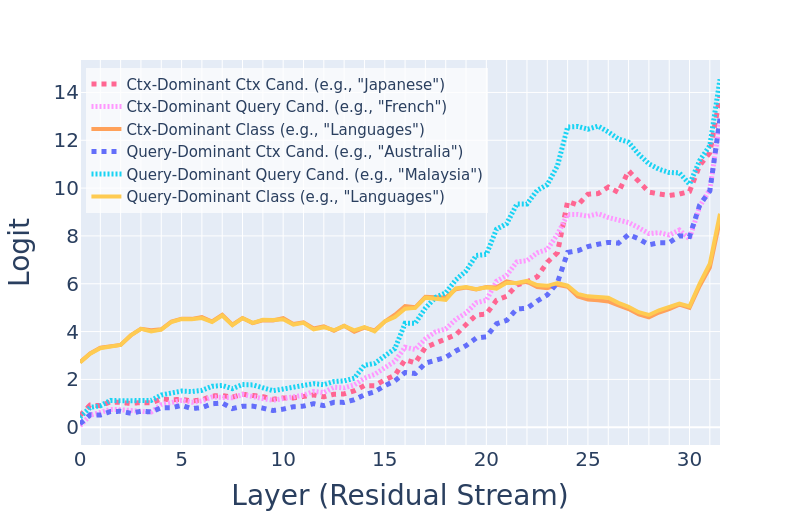}
        \caption{\textbf{C+Q}: Token logit from accumulated residual stream. ($R_{T,l}^{\text{1}}$, $R_{T,l}^{\text{2}}$) are visualized per layer.}
        \label{fig:c+q token logit llama - appendix}
    \end{subfigure}
    \vspace{0.5em} 
    \begin{subfigure}[b]{0.48\textwidth}
        \centering
        \includegraphics[width=\textwidth]{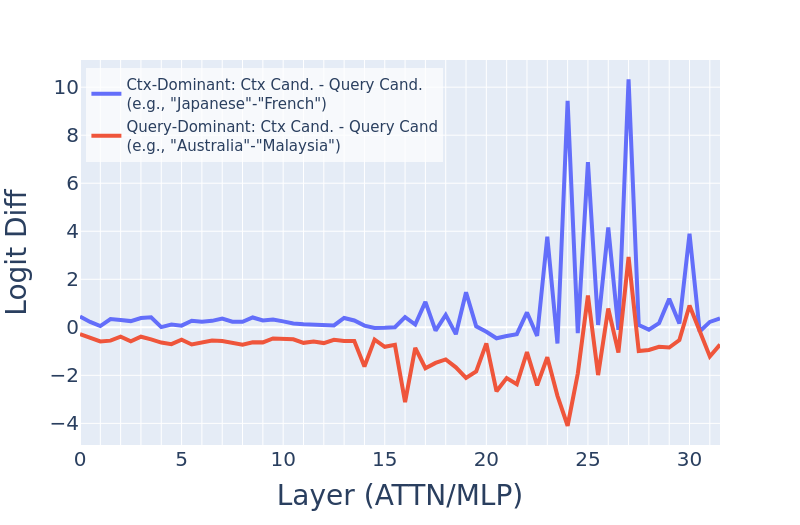}
        \caption{\textbf{C+Q}: candidate logit difference from attention and MLP output. $A_{T,l}$ and $M_{T,l}$ are visualized per layer.}
        \label{fig:c+q answer logit diff llama}
    \end{subfigure}
    \caption{Additional logit attribution results for \textbf{Llama-3 8B}.}
    \label{fig: logit-attribution-additional-results-llama}
    
    
    %
\end{figure*}

Additional results for Llama 8B are presented in Figure \ref{fig: logit-attribution-additional-results-llama}. Importantly, we point out that the class-based generalization might have existed already for the Q-only case. In Figure \ref{fig:q-only token logit llama}, we observe a similar pattern as the C+Q case presented in Figure \ref{fig: c+q token logit} -- models build abstract class representation in the lower layers, before refining their answers to concrete ones. In fact, when we plot the logit difference of the abstract class tokens under C+Q and Q-only case in Figure \ref{fig:c+q-q logit diff llama}, as shown as orange and yellow lines for context-dominant and query-dominant case, the lines center around 0 -- suggesting that the computation of abstract class representations exists for zero-shot case, and is not influenced by the added irrelevant context.

\begin{figure*}[h!]
    \centering
    \begin{subfigure}[b]{0.48\textwidth} 
        \centering
        \includegraphics[width=\textwidth]{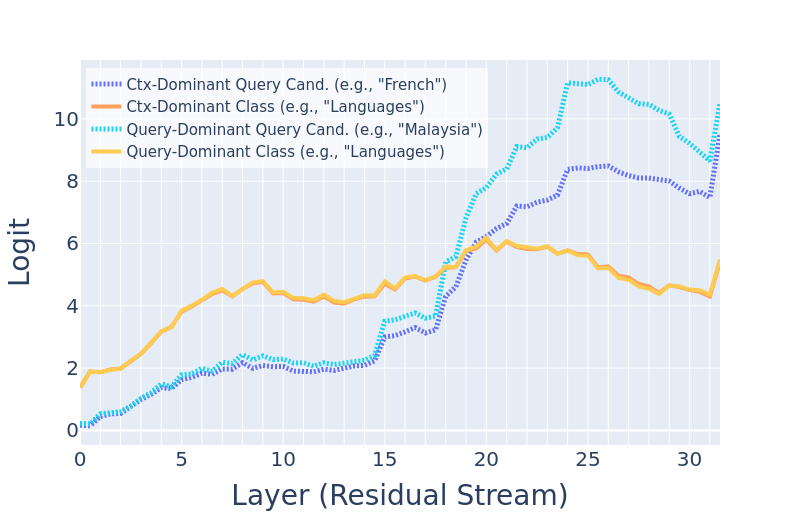}
        \caption{\textbf{Q-only:} Token logit from accumulated residual stream. ($R_{T,l}^{\text{1}}$, $R_{T,l}^{\text{2}}$) are visualized per layer.}
        \label{fig:q-only token logit mistral}
    \end{subfigure}
    \vspace{0.5em} 
    \begin{subfigure}[b]{0.48\textwidth}
        \centering
        \includegraphics[width=\textwidth]{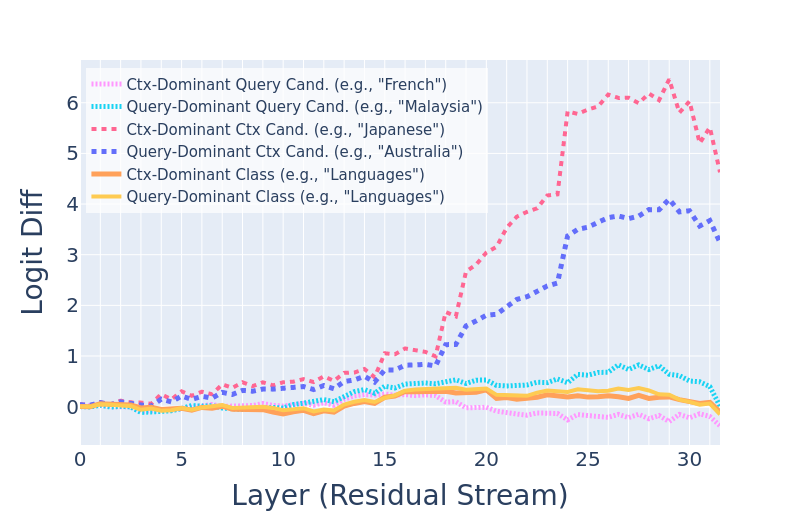}
        \caption{Token logit difference (Logit in \textbf{C+Q} - Logit in \textbf{Q-only}) from accumulated residual stream ($R_{T,l}^{\text{1}}$, $R_{T,l}^{\text{2}}$).}
        \label{fig:c+q-q logit diff mistral}
    \end{subfigure}
    \begin{subfigure}[b]{0.48\textwidth} 
        \centering
        \includegraphics[width=\textwidth]{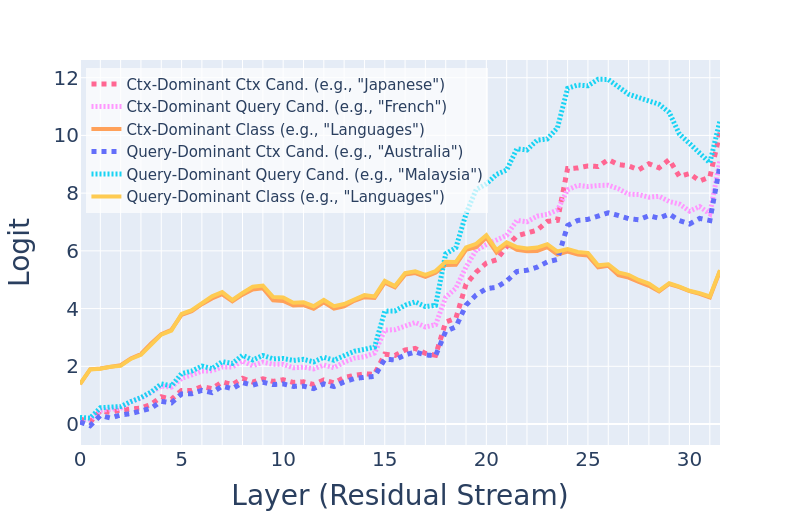}
        \caption{\textbf{C+Q}: Token logit from accumulated residual stream. ($R_{T,l}^{\text{1}}$, $R_{T,l}^{\text{2}}$) are visualized per layer.}
        \label{fig:c+q token logit mistral - appendix}
    \end{subfigure}
    \vspace{0.5em} 
    \begin{subfigure}[b]{0.48\textwidth}
        \centering
        \includegraphics[width=\textwidth]{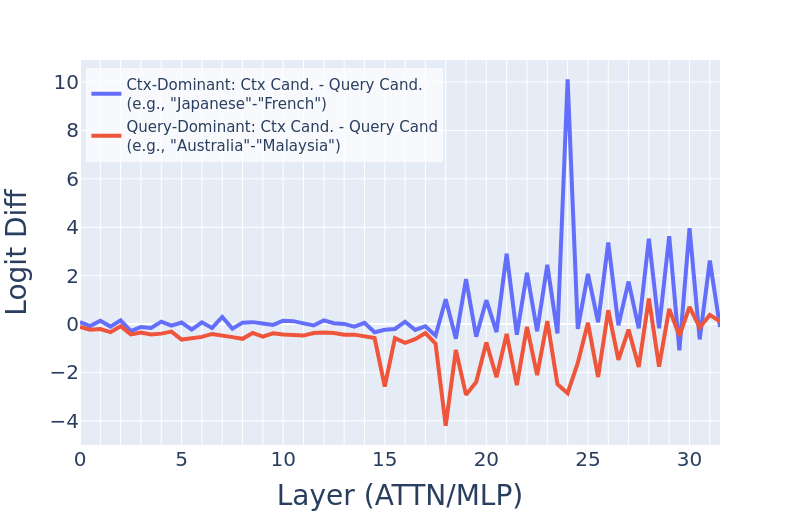}
        \caption{\textbf{C+Q}: candidate logit difference from attention and MLP output. $A_{T,l}$ and $M_{T,l}$ are visualized per layer.}
        \label{fig:c+q answer logit diff mistral}
    \end{subfigure}
    
    \caption{Logit Attribution Results For \textbf{Mistral 7B}.}
    \label{fig: logit-attribution-additional-results-mistral}
\end{figure*}

\begin{figure*}[h!]
    \centering
    \begin{subfigure}[b]{0.48\textwidth} 
        \centering
        \includegraphics[width=\textwidth]{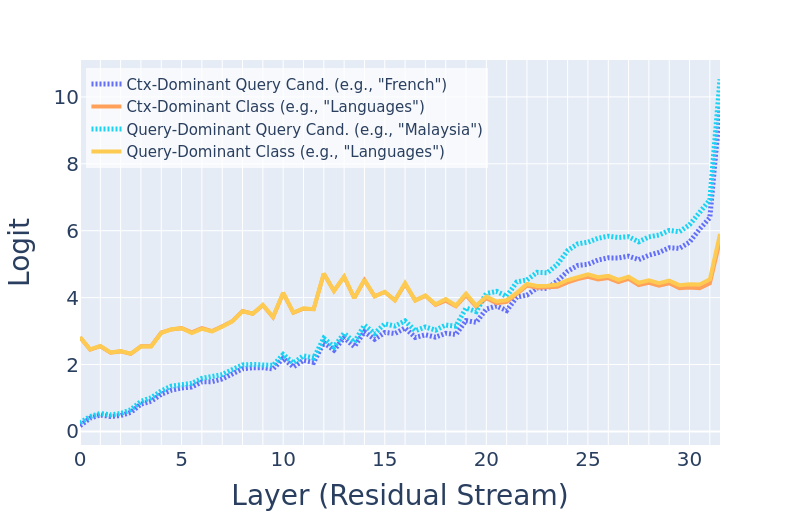}
        \caption{\textbf{Q-only:} Token logit from accumulated residual stream. ($R_{T,l}^{\text{1}}$, $R_{T,l}^{\text{2}}$) are visualized per layer.}
        \label{fig:q-only token logit pythia}
    \end{subfigure}
    \vspace{0.5em} 
    \begin{subfigure}[b]{0.48\textwidth}
        \centering
        \includegraphics[width=\textwidth]{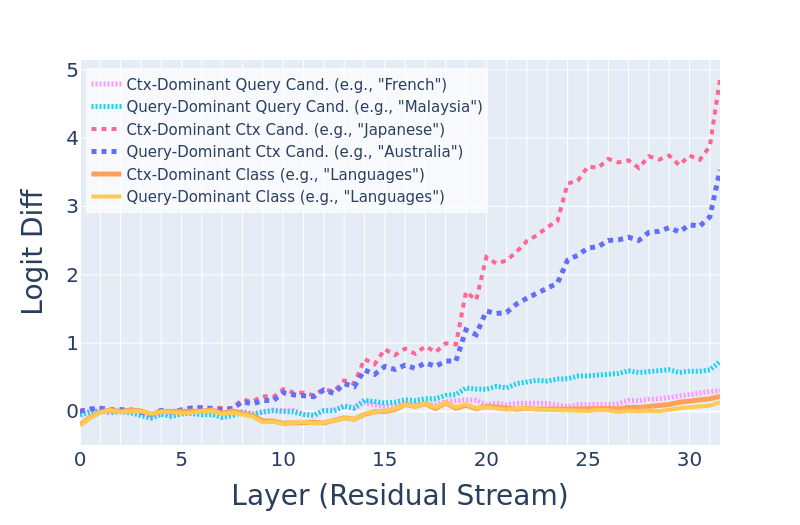}
        \caption{Token logit difference (Logit in \textbf{C+Q} - Logit in \textbf{Q-only}) from accumulated residual stream ($R_{T,l}^{\text{1}}$, $R_{T,l}^{\text{2}}$).}
        \label{fig:c+q-q logit diff pythia}
    \end{subfigure}
    \begin{subfigure}[b]{0.48\textwidth} 
        \centering
        \includegraphics[width=\textwidth]{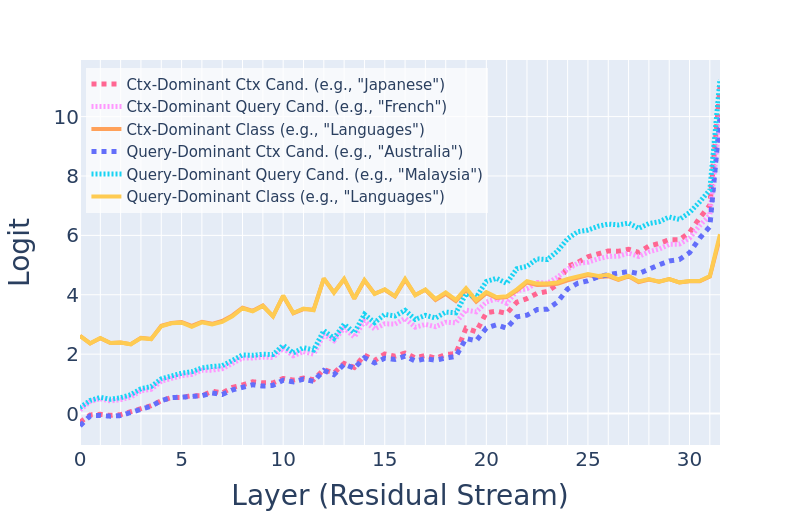}
        \caption{\textbf{C+Q}: Token logit from accumulated residual stream. ($R_{T,l}^{\text{1}}$, $R_{T,l}^{\text{2}}$) are visualized per layer.}
        \label{fig:c+q token logit pythia}
    \end{subfigure}
    \vspace{0.5em} 
    \begin{subfigure}[b]{0.48\textwidth}
        \centering
        \includegraphics[width=\textwidth]{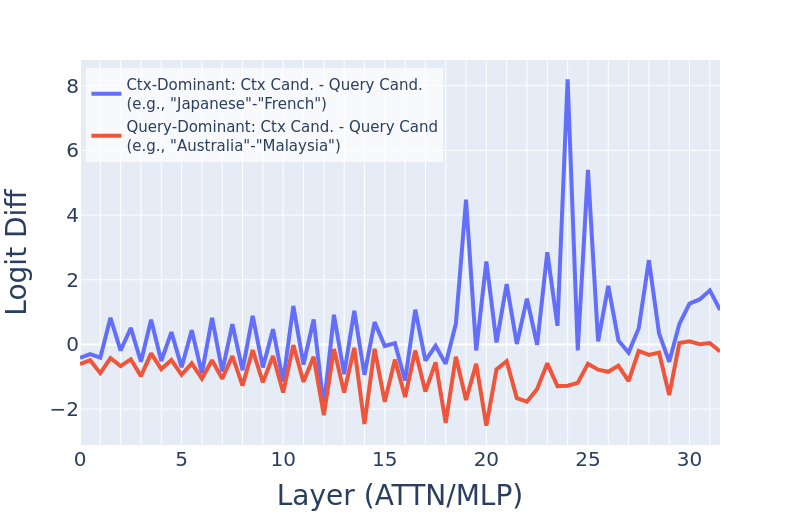}
        \caption{\textbf{C+Q}: candidate logit difference from attention and MLP output. $A_{T,l}$ and $M_{T,l}$ are visualized per layer.}
        \label{fig:c+q answer logit diff pythia}
    \end{subfigure}
    
    \caption{Logit Attribution Results For \textbf{Pythia 6.9B}.}
    \label{fig: logit-attribution-additional-results-pythia}
\end{figure*}

Logit attribution results for Mistral 7B are presented in Figure \ref{fig: logit-attribution-additional-results-mistral}, and for Pythia 6.9B in Figure \ref{fig: logit-attribution-additional-results-pythia}. We remark that these plots follow a similar pattern.

\section{Activation Patching} \label{appendix: activation-patching}
\subsection{Implementation Details}
In the corrupted run, we corrupt the embeddings of all tokens for context subject and object in context patching, and all tokens for query subject in query patching by adding a Gaussian noise where $\sigma$ is 3 times of the empirical standard deviation of the input embeddings over a body of text ($sigma \approx 0.3)$ \cite{meng2022locating}.

\subsection{Additional Activation Patching Results}
\begin{figure*}[htbp]

    \centering
    \begin{subfigure}{\linewidth}
        \centering
        \includegraphics[width=\linewidth]{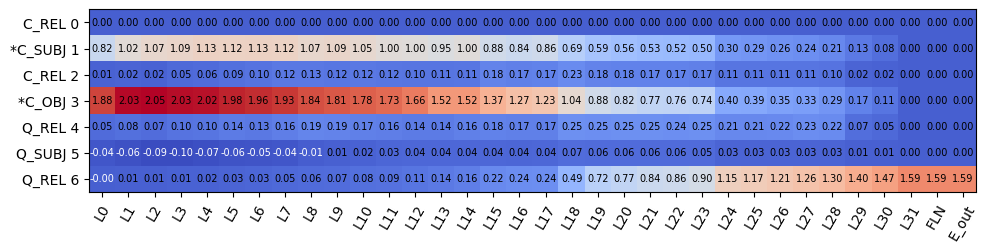}
        \caption{C+Q: Context circuit in context-dominant case.}
        \label{fig: context-dominant-context-circuit-mistral}
    \end{subfigure}
    
    \begin{subfigure}{\linewidth}
        \centering
        \includegraphics[width=\linewidth]{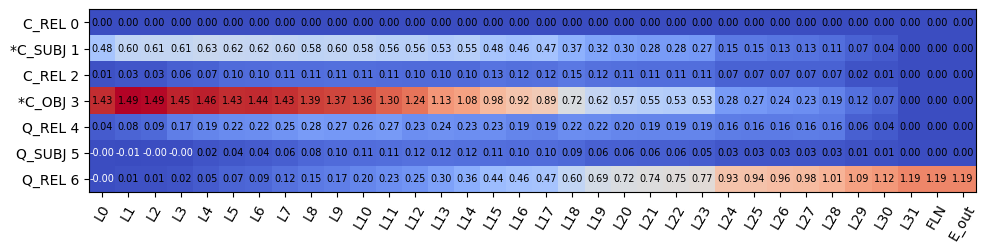}
        \caption{C+Q: Context circuit in query-dominant case.}
        \label{fig: query-dominant-ctx-circuit-mistral}
    \end{subfigure}
    
    \begin{subfigure}{\linewidth}
        \centering
        \includegraphics[width=\linewidth]{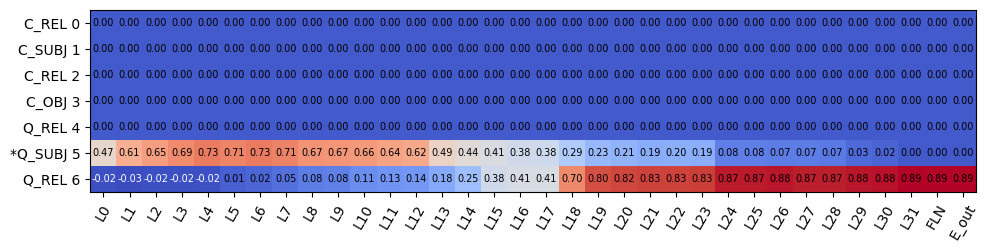}
        \caption{C+Q: Query circuit in context-dominant case.}
        \label{fig: ctx-driven-query-circuit-mistral}
    \end{subfigure}
    
    \begin{subfigure}{\linewidth}
        \centering
        \includegraphics[width=\linewidth]{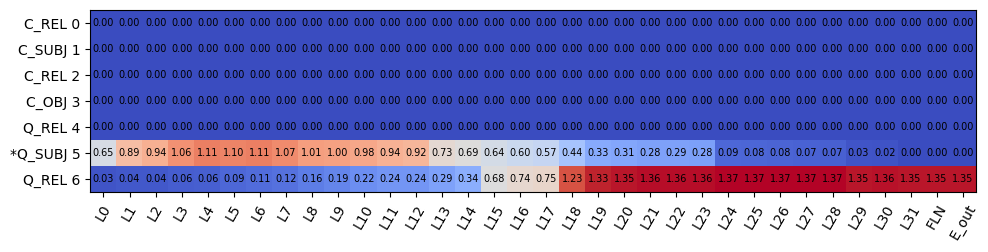}
        \caption{C+Q: Query circuit in query-dominant case.}
        \label{fig: query-dominant-query-circuit-mistral}
    \end{subfigure}
    
    \caption{Activation patching under C+Q condition for \textbf{Mistral 7B}. \label{fig: activation-patching-mistral}}
\end{figure*}

\begin{figure*}[htbp]

    \centering
    \begin{subfigure}{\linewidth}
        \centering
        \includegraphics[width=\linewidth]{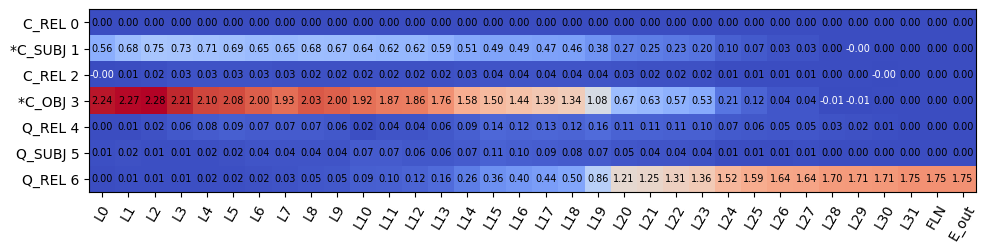}
        \caption{Context circuit in context-dominant case.}
        \label{fig: context-dominant-context-circuit-pythia}
    \end{subfigure}
    
    \begin{subfigure}{\linewidth}
        \centering
        \includegraphics[width=\linewidth]{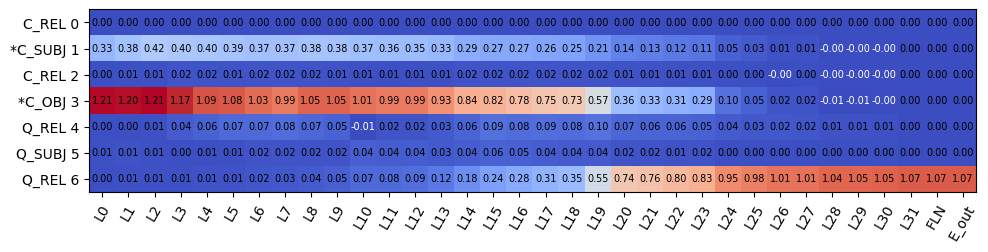}
        \caption{Context circuit in query-dominant case.}
        \label{fig: query-dominant-ctx-circuit-pythia}
    \end{subfigure}
    
    \begin{subfigure}{\linewidth}
        \centering
        \includegraphics[width=\linewidth]{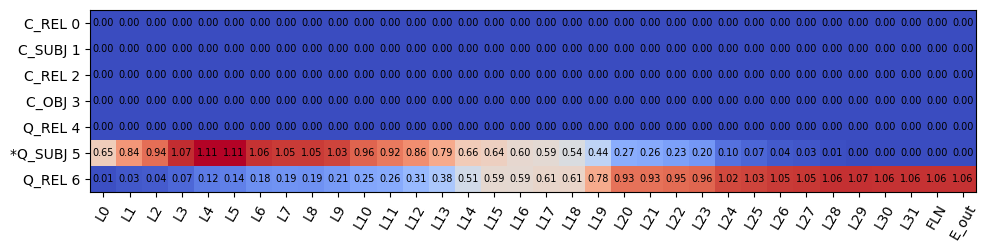}
        \caption{Query circuit in context-dominant case.}
        \label{fig: ctx-driven-query-circuit-pythia}
    \end{subfigure}
    
    \begin{subfigure}{\linewidth}
        \centering
        \includegraphics[width=\linewidth]{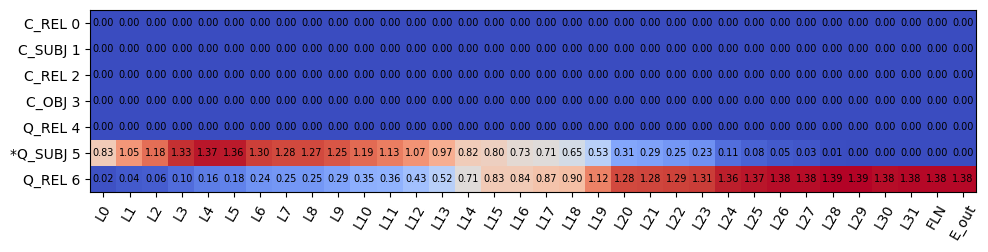}
        \caption{Query circuit in query-dominant case.}
        \label{fig: query-dominant-query-circuit-pythia}
    \end{subfigure}
    
    \caption{Activation patching under C+Q condition for \textbf{Pythia 6.9B}.\label{fig: activation-patching-pythia}}
\end{figure*}

Activation patching results under the C+Q condtion for Mistral and Pythia are in Figure~\ref{fig: activation-patching-mistral} and~\ref{fig: activation-patching-pythia}, respectively.

\begin{figure*}[htbp]
    \centering
    \begin{subfigure}{\linewidth}
        \centering
        \includegraphics[width=\linewidth]{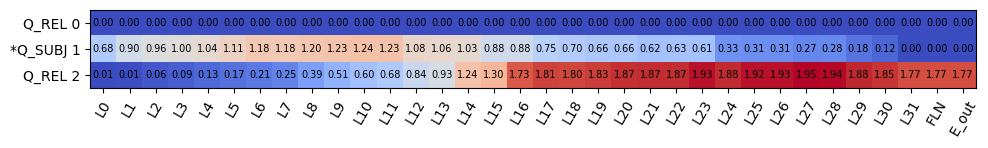}
        \caption{Q-only: Query circuit in context-dominant case.}
        \label{fig: Q-only-context-dominant-query-circuit-llama}
    \end{subfigure}
    
    \begin{subfigure}{\linewidth}
        \centering
        \includegraphics[width=\linewidth]{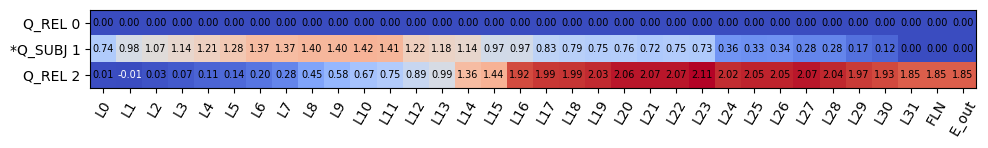}
        \caption{Q-only: Query circuit in query-dominant case.}
        \label{fig: Q-only-query-dominant-query-circuit-llama}
    \end{subfigure}
    
    
    
    \caption{Activation patching under Q-Only condition reveals that query circuit is the same before and after adding the irrelevant context for \textbf{Llama-3 8B}.\label{fig: Q-only-query-circuit-llama}}
\end{figure*}

\begin{figure*}[htbp]
    \centering
    \begin{subfigure}{\linewidth}
        \centering
        \includegraphics[width=\linewidth]{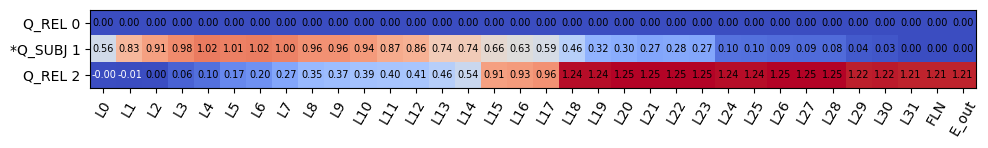}
        \caption{Q-only: Query circuit in context-dominant case.}
        \label{fig: Q-only-context-dominant-query-circuit-mistral}
    \end{subfigure}
    
    \begin{subfigure}{\linewidth}
        \centering
        \includegraphics[width=\linewidth]{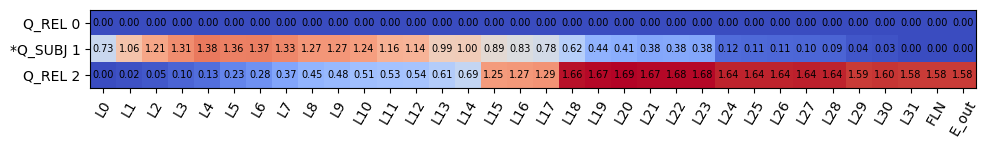}
        \caption{Q-only: Query circuit in query-dominant case.}
        \label{fig: Q-only-query-dominant-query-circuit-mistral}
    \end{subfigure}
        \caption{Activation patching under Q-Only condition reveals that query circuit is the same before and after adding the irrelevant context for \textbf{{Mistral 7B}}.\label{fig: Q-only-query-circuit-mistral}}
    
\end{figure*}

\begin{figure*}[htbp]
    \centering
    \begin{subfigure}{\linewidth}
        \centering
        \includegraphics[width=\linewidth]{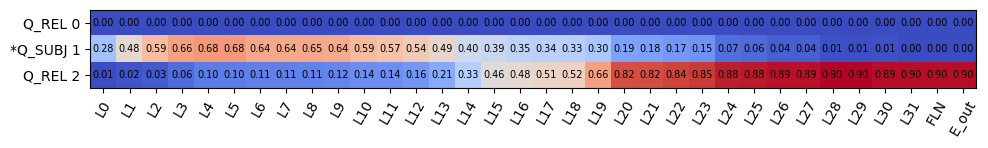}
        \caption{Q-only: Query Circuit in context-dominant case.}
        \label{fig: Q-only-context-dominant-query-circuit-pythia}
    \end{subfigure}
    
    \begin{subfigure}{\linewidth}
        \centering
        \includegraphics[width=\linewidth]{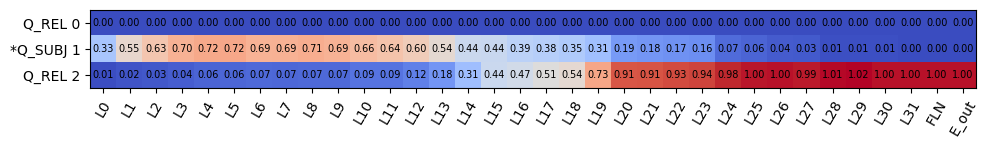}
        \caption{Q-only: Query Circuit in query-dominant case.}
        \label{fig: Q-only-query-dominant-query-circuit-pythia}
    \end{subfigure}
    \caption{Activation patching under Q-Only condition reveals that query circuit is the same before and after adding the irrelevant context for \textbf{{Pythia 6.9B}}.\label{fig: Q-only-query-circuit-pythia}}

\end{figure*}

Additionally, we also visualize the query circuit under the Q-only condition in Figure~\ref{fig: Q-only-query-circuit-llama}, ~\ref{fig: Q-only-query-circuit-mistral}, and ~\ref{fig: Q-only-query-circuit-pythia}, for Llama, Mistral, and Pythia, respectively. We remark on two important observations: (i) The query circuit is the same for context-dominant and query-dominant data, without irrelevant context. (ii) The query circuit remains as is after adding the irrelevant context, as compared to Figures \ref{fig: ctx-driven-query-circuit-llama} and \ref{fig: query-dominant-query-circuit-llama}.



\section{Attention Knockout} \label{appendix: attention knockout}
\subsection{Implementation Details}
In the attention knockout experiments, our goal is to see if we can intervene in the internal computation to change the output behavior. Specifically, in context-dominant case, we would like to flip the prediction $A_{C+Q}$ from $C_{\text{cand.}}$ (e.g., `Japanese' to $Q_{\text{cand.}}$ `French'; And in query-dominant case, we would like to flip the prediction $A_{C+Q}$ from $Q_{\text{cand.}}$ `Malaysia' to $C_{\text{cand.}}$ `Australia'. 

To do this, we intervene in two layers: the first attention layer where the context information is transferred to the last token residual stream, and the attention layer where the most context information is written into the last token residual stream. These two layers correspond to the first blue spike and the highest blue spike in Figures~\ref{fig:c+q answer logit diff llama}, ~\ref{fig:c+q answer logit diff mistral} and ~\ref{fig:c+q answer logit diff pythia}. For Llama-3, it is layers 17 and 24, respectively. For Mistral, it is layers 18 and 24, respectively. For Pythia, it is layers 19 and 24, respectively.

Specifically, in the context-dominant case, at the last token position, we set the attention scores corresponding to all tokens in the context to be $-\infty$, therefore, attention weight (which sums up to 1) is only a distribution over the query tokens. We perform this intervention to block information flow from the context to the last token position, and we only allow models to attend to the query part. Similarly, in the query-dominant case, we set the attention scores corresponding to all tokens in the query to be $-\infty$, allowing the models to only retrieve information from the context.

To compare the knockout effect of the two critical layers with other layers, we select two random lower layers and two random higher layers. We report the average intervention results for three runs.

\subsection{Additional Results}
Results for Llama and Mistral are presented in Table~\ref{tab:knockout_results_llama_detailed} and Table~\ref{tab:knockout_results_mistral}, respectively.

With the targeted two-critical-layer intervention:
\begin{itemize}
    \item Llama: 465/1000 context-dominant datapoints flip to query-based candidates, while 407/1000 remain context-based. Conversely, 225/1000 query-dominant datapoints shift to context-based candidates, while 704/1000 remain query-based.
    \item Mistral: 437/1000 context-dominant datapoints flip to query-based candidates, while 514/1000 remain context-based. Similarly, 232/1000 query-dominant datapoints shift to context-based candidates, while 713/1000 remain query-based.
    \item Pythia: 470/1000 context-dominant datapoints flip to query-based candidates, while 486/1000 remain context-based. Conversely, 294/1000 query-dominant datapoints shift to context-based candidates, while 648/1000 remain query-based.
\end{itemize}
Across all models, approximately 950 datapoints remain context- or query-based candidates, instead of random non-identifiable answers, indicating that our intervention preserves model capabilities.

\begin{table}[h]
    \centering
    \small
    \setlength{\tabcolsep}{4pt} 
    \renewcommand{\arraystretch}{0.8} 
    \begin{tabular}{lcclcccc}
        \toprule
        & Orig. & \multicolumn{2}{c}{L17+L24} & \multicolumn{2}{c}{2 Low} & \multicolumn{2}{c}{2 High} \\
        \cmidrule(lr){2-2} \cmidrule(lr){3-4} \cmidrule(lr){5-6} \cmidrule(lr){7-8}
        & Prob. & Prob. & $\Delta$ & Prob. & $\Delta$ & Prob. & $\Delta$ \\
        \midrule
        \multicolumn{8}{c}{\textbf{Context-Dominant}} \\
        \midrule
        Ctx   & 22.6 & \textbf{14.6} & -8.0 & 19.9 & -2.7 & 19.0 & -3.6 \\
        Query & 8.2  & \textbf{12.4} & +4.2  & 8.2  & +0.0  & 9.2  & +1.0  \\
        \midrule
        \multicolumn{8}{c}{\textbf{Query-Dominant}} \\
        \midrule
        Query & 33.0 & \textbf{25.0} & -8.0 & 25.5 & -7.5 & 31.7 & -1.3 \\
        Ctx   & 6.5  & \textbf{10.7} & +4.2  & 7.7  & +1.2  & 6.4  & -0.1 \\
        \bottomrule
    \end{tabular}
    \caption{Effect of attention knockout on context- (Ctx) and query-based (Query) candidate probabilities on Llama-3. ``Orig.'' = No intervention, “2 Low” = Two lower layers (<17), “2 High” = Two higher layers (>24). “Diff.” represents the probability difference, and $\Delta$ denotes the change from the original setting. \textbf{(Mistral 7B)}}
    \label{tab:knockout_results_mistral}
\end{table}

\begin{table}[h]
    \centering
    \small
    \setlength{\tabcolsep}{4pt} 
    \renewcommand{\arraystretch}{0.8} 
    \begin{tabular}{lcclcccc}
        \toprule
        & Orig. & \multicolumn{2}{c}{L17+L24} & \multicolumn{2}{c}{2 Low} & \multicolumn{2}{c}{2 High} \\
        \cmidrule(lr){2-2} \cmidrule(lr){3-4} \cmidrule(lr){5-6} \cmidrule(lr){7-8}
        & Prob. & Prob. & $\Delta$ & Prob. & $\Delta$ & Prob. & $\Delta$ \\
        \midrule
        \multicolumn{8}{c}{\textbf{Context-Dominant}} \\
        \midrule
        Ctx   & 25.5 & \textbf{13.1} & -12.4 & 20.9 & -4.6 & 21.1 & -4.4 \\
        Query & 8.6  & \textbf{14.8} & +6.2 & 8.9 & +0.3 & 12.6 & +4.0 \\

        \midrule
        \multicolumn{8}{c}{\textbf{Query-Dominant}} \\
        \midrule
        Query & 35.2 & \textbf{26.8} & -8.4  & 25.7 & -9.5 & 33.4 & -1.8 \\
        Ctx   & 6.6  & \textbf{11.3} & +4.7  & 7.7  & +1.1 & 7.1  & +0.5 \\

        \bottomrule
    \end{tabular}
    \caption{Effect of attention knockout on context- (Ctx) and query-based (Query) candidate probabilities on Llama-3. ``Orig.'' = No intervention, “2 Low” = Two lower layers (<17), “2 High” = Two higher layers (>24). “Diff.” represents the probability difference, and $\Delta$ denotes the change from the original setting. \textbf{(Llama-3)}}
    \label{tab:knockout_results_llama_detailed}
\end{table}

\begin{table}[h]
    \centering
    \small
    \setlength{\tabcolsep}{4pt} 
    \renewcommand{\arraystretch}{0.8} 
    \begin{tabular}{lcclcccc}
        \toprule
        & Orig. & \multicolumn{2}{c}{L17+L24} & \multicolumn{2}{c}{2 Low} & \multicolumn{2}{c}{2 High} \\
        \cmidrule(lr){2-2} \cmidrule(lr){3-4} \cmidrule(lr){5-6} \cmidrule(lr){7-8}
        & Prob. & Prob. & $\Delta$ & Prob. & $\Delta$ & Prob. & $\Delta$ \\
        \midrule
        \multicolumn{8}{c}{\textbf{Context-Dominant}} \\
        \midrule
        Ctx   & 22.3 & \textbf{13.6} & -8.7 & 18.3 & -4.0 & 20.5 & -1.8 \\
        Query & 7.4  & \textbf{11.5} & +4.1 & 8.4 & +1.0 & 7.8 & +0.4 \\

        \midrule
        \multicolumn{8}{c}{\textbf{Query-Dominant}} \\
        \midrule
        Query & 26.6 & \textbf{20.8} & -5.8 & 20.5 & -6.1 & 25.6 & -1.0 \\
        Ctx   &  6.3 & \textbf{9.6} & +3.3 & 6.9 & +0.6 & 6.2 & -0.1 \\

        \bottomrule
    \end{tabular}
    \caption{Effect of attention knockout on context- (Ctx) and query-based (Query) candidate probabilities on Llama-3. ``Orig.'' = No intervention, “2 Low” = Two lower layers (<17), “2 High” = Two higher layers (>24). “Diff.” represents the probability difference, and $\Delta$ denotes the change from the original setting. \textbf{(Pythia)}}
    \label{tab:knockout_results_pythia_detailed}
\end{table}

\section{Ablation Studies}
\subsection{Experimental Details}\label{app: larger-models-experiment-details} 
Due to computational constraints, we cannot inference on the full $C+Q$ dataset with 102M for larger-sized model, we therefore conduct sampling as follows: For Pythia-12B, we first randomly sample 1,000 datapoints from the ParaRel dataset, then randomly sample 100 datapoints per relation for the context demonstrations, resulting in around 3.9M datapoints. For Llama-70B, we first randomly sample 500 datapoints from the ParaRel dataset, then randomly sample 50 datapoints per relation for the context demonstrations, resulting in around 975K datapoints.

\subsection{Statistical Validation of Contextual Influence}
\begin{table}[h!]
\centering
\small
\begin{tabular}{@{}lccc@{}}
\toprule
\textbf{Value} & \textbf{Pythia 12B} & \textbf{Llama-3 70B}  & \textbf{Llama-3 8B Prompt} \\
\midrule
Mean PMI       & 4.19                & 4.22           & 4.2756      \\
T-statistics   & 11.74               & 11.89   &13.0924             \\
\textit{p}-value     & 0.0002              & 0.0001  &0.0001             \\
\bottomrule
\end{tabular}
\caption{Statistical analysis of PMI between context terms and generated context-based candidates.}
\label{tab:statistical-pmi}
\end{table}


\newpage

\begin{center}
\onecolumn
\begin{longtable}{|c|p{6cm}|c|c|}
\hline
\textbf{Relation} & \textbf{Template} & \textbf{Ctx Type} & \textbf{Total Rows} \\
\hline
\endfirsthead

\hline
\textbf{Relation} & \textbf{Template} & \textbf{Type} & \textbf{Total Rows} \\
\hline
\endhead

P1001 & [X] is a legal term in [Y] & Place & 664 \\
P101 & The expertise of [X] is [Y]. & Others & 571 \\
P103 & The mother tongue of [X] is [Y]. & Language & 919 \\
P106 & [X] works as [Y]. & Job & 821 \\
P108 & [X], who is employed by [Y]. & Company & 378 \\
P127 & [X] owner [Y]. & Company & 616 \\
P1303 & [X] plays the [Y]. & Others & 513 \\
P131 & [X] is in [Y]. & Place & 775 \\
P136 & [X] plays [Y]. & Others & 859 \\
P1376 & [X], the capital city of [Y]. & Place & 179 \\
P138 & [X], which is named after [Y]. & Others & 461 \\
P140 & [X] is follower of [Y]. & Others & 432 \\
P1412 & [X] communicated in [Y]. & Language & 924 \\
P159 & [X] is headquartered in [Y]. & Place & 801 \\
P17 & [X], located in [Y]. & Place & 912 \\
P176 & [X], produced by [Y]. & Company & 925 \\
P178 & [X], a product developed by [Y]. & Company & 588 \\
P19 & [X] is native to [Y]. & Place & 779 \\
P190 & [X] is a twin city of [Y]. & Place & 671 \\
P20 & [X] passed away at [Y]. & Place & 817 \\
P264 & [X]'s label is [Y]. & Company & 53 \\
P27 & [X], a citizen of [Y]. & Place & 958 \\
P276 & [X] is located in [Y]. & Place & 764 \\
P279 & [X], a type of [Y]. & Others & 900 \\
P30 & [X] is a part of the continent of [Y]. & Place& 959 \\
P36 & The capital city of [X] is [Y]. & Place & 471 \\
P361 & [X] is a part of [Y]. & Others & 746 \\
P364 & The original language of [X] was [Y]. & Language & 756 \\
P37 & The official language of [X] is [Y]. & Language & 900 \\
P39 & [X], who holds the position of [Y]. & Job & 485 \\
P407 & [X] was written in [Y]. & Language & 857 \\
P413 & [X] plays in the position of [Y]. & Job & 952 \\
P449 & [X] premiered on [Y]. & Company & 801 \\
P463 & [X] belongs to the organization of [Y]. & Company & 203 \\
P47 & [X] borders with [Y]. & Place & 649 \\
P495 & [X] was formed in [Y]. & Place & 905 \\
P530 & [X] ties diplomatic relations with [Y]. & Place & 950 \\
P740 & [X], founded in [Y]. & Place & 843 \\
P937 & [X] found employment in [Y]. & Place & 853 \\
\hline
\caption{Overview of Relations, Templates, Types, and Total Rows in the original Pararel Dataset. We take this dataset and construct the $C+Q$ dataset, which has around 106.6M rows.}
\label{tab: pararel_dataset}
\end{longtable}
\end{center}

\begin{center}
\onecolumn
\begin{longtable}{|c|p{6cm}|c|c|}
\hline
\textbf{Relation} & \textbf{Template} & \textbf{Ctx Type} & \textbf{Total Rows} \\
\hline
\endfirsthead

\hline
\textbf{Relation} & \textbf{Template} & \textbf{Ctx Type} & \textbf{Total Rows} \\
\hline
\endhead

P1001 & [X] is a legal term in [Y]. & Place & 664 \\
P101 & [X]'s expertise is [Y]. & Others & 571 \\
P103 & The mother tongue of [X] is [Y]. & Language & 919 \\
P106 & The occupation of [X] is [Y]. & Job & 821 \\
P108 & [X] works for [Y]. & Company & 378 \\
P127 & [X] is owned by [Y]. & Company & 616 \\
P1303 & [X] plays [Y]. & Others & 513 \\
P131 & [X] is in [Y]. & Place & 775 \\
P136 & [X], who plays [Y]. & Others & 859 \\
P1376 & [X], the capital city of [Y]. & Place & 179 \\
P138 & [X] is called after [Y]. & Others & 461 \\
P140 & [X] is follower of [Y]. & Others & 432 \\
P1412 & [X] communicated in [Y]. & Language & 924 \\
P159 & The headquarters of [X] is in [Y]. & Place & 801 \\
P17 & [X], which is located in [Y]. & Place & 912 \\
P176 & [X], developed by [Y]. & Company & 925 \\
P178 & [X], created by [Y]. & Company & 588 \\
P19 & [X] originates from [Y]. & Place & 779 \\
P190 & [X] is a twin city of [Y]. & Place & 671 \\
P20 & [X] died in [Y]. & Place & 817 \\
P264 & [X], which is represented by [Y]. & Company & 53 \\
P27 & [X] has a citizenship of [Y]. & Place & 958 \\
P276 & [X] is in [Y]. & Place & 764 \\
P279 & [X],  a type of [Y]. & Others & 900 \\
P30 & [X] belongs to the continent of [Y]. & Place & 959 \\
P36 & The capital of [X] is [Y]. & Place & 471 \\
P361 & [X] is part of [Y]. & Others & 746 \\
P364 & The original language of [X] was [Y]. & Language & 756 \\
P37 & The official language of [X] is [Y]. & Language & 900 \\
P39 & [X], whose position is that of [Y]. & Job & 485 \\
P407 & The language of [X] is [Y]. & Language & 857 \\
P413 & [X] plays in the position of [Y]. & Job & 952 \\
P449 & [X] debuted on [Y]. & Company & 801 \\
P463 & [X] is a member of [Y]. & Company & 203 \\
P47 & [X] borders with [Y]. & Place & 649 \\
P495 & [X] formed in [Y]. & Place & 905 \\
P530 & [X] maintains diplomatic relations with [Y]. & Place & 950 \\
P740 & [X], founded in [Y]. & Place & 843 \\
P937 & [X] was employed in [Y]. & Place & 853 \\
\hline
\caption{Overview of relations, alternative prompt templates, types, and total rows in the original ParaRel dataset. We use these randomly sampled alternative templates to test sensitivity to prompt phrasing.}
\label{tab: pararel_dataset_template_variation}
\end{longtable}
\end{center}

\newpage

\renewcommand{\arraystretch}{1.3}
\begin{longtable}{p{6cm} p{8cm}}
    \toprule
    \textbf{Category} & \textbf{Details} \\
    \midrule
    \endfirsthead

    \toprule
    \textbf{Category} & \textbf{Details} \\
    \midrule
    \endhead

    \multicolumn{2}{l}{\textbf{Example 1}} \\
    \midrule
    Context & \textbf{Hanoi} is a twin city of \textbf{Bangkok}. \\
    Query & The \textbf{mother tongue} of Louis Legendre is \\
    Class & Languages \\
    \midrule
    Context Subject Possible Answers & Vietnamese, Tay, Hmong, Khmer, English, French, Chinese \\
    Context Object Possible Answers & Thai, Lao, Chinese, Malay, Khmer \\
    Context-Based Candidates & Vietnamese, Thai \\
    \midrule
    Context-Influenced? & True \\
    Correct Class? & True\\
    Exists Answer that Satisfies Both? & True \\
    \midrule

    \multicolumn{2}{l}{\textbf{Example 2}} \\
    \midrule
    Context & \textbf{Svend Asmussen} plays the \textbf{violin}. \\
    Query & Social-Economic Council is a legal term \textbf{in} \\
    Class & Places (Countries, Cities, States, etc.)/Languages\\
    \midrule
    Context Subject Possible Answers & Danmark, Danish (Svend Asmussen is a Violinist and jazz musician) \\
    Context Object Possible Answers & Italy, Italian (Violin was originated in Italy) \\
    Context-Based Candidates & Denmark \\
    \midrule
    Context-Influenced? & True \\
    Correct Class? & True\\
    Exists Answer that Satisfies Both? & True \\
    \midrule
    \multicolumn{2}{l}{\textbf{Example 3}} \\
    \midrule
    Context & \textbf{Manchester Business School} is headquartered in \textbf{Manchester}. \\
    Query & Antipope Paschal III, who holds the \textbf{position} of \\
    Class & Jobs/Positions/Roles \\
    \midrule
    Context Subject Possible Answers & Professor, Lecturer, Instructor, Researcher, Department Chair, Provost, Dean, Academic Advisor, Teaching Assistant, Student, etc. \\
    Context Object Possible Answers & N/A \\
    Context-Based Candidates & Dean, Professor \\
    \midrule
    Context-Influenced? & True \\
    Correct Class? & True\\
    Exists Answer that Satisfies Both? & True \\
    \midrule
    \multicolumn{2}{l}{\textbf{Example 4}} \\
    \midrule
    Context & \textbf{Saharsa district} is in \textbf{Bihar}. \\
    Query & Colbert Mountains is a part of the \textbf{continent} of \\
    Class & Continents/Places\\
    \midrule
    Context Subject Possible Answers & Asia \\
    Context Object Possible Answers & Asia \\
    Context-Based Candidates & Asia, India \\
    \midrule
    Context-Influenced? & True \\
    Correct Class? & True\\
    Exists Answer that Satisfies Both? & True \\

    \midrule
    \multicolumn{2}{l}{\textbf{Example 5}} \\
    \midrule
    Context & \textbf{Lee Jong-hyun} plays the \textbf{guitar}. \\
    Query & Northern Foothills is a part of the \textbf{continent} of \\
    Class & Continents \\
    \midrule
    Context Subject Possible Answers & Asia \\
    Context Object Possible Answers & Europe (Guitar originated in Spain) \\
    Context-Based Candidates & South, Korea \\
    \midrule
    Context-Influenced? & True \\
    Correct Class? & False\\
    Exists Answer that Satisfies Both? & False \\
    
    \midrule
    \multicolumn{2}{l}{\textbf{Example 6}} \\
    \midrule
    Context & Grant Green plays jazz. \\
    Query & David Gates plays the \\
    Class & Role/Genre/Style/Position/Musical Instrument \\
    \midrule
    Context Subject Possible Answers &  guitarist, composer, musician, songwriter etc. (role of Grant Green), guitar (Musical Instrument that Grant Green plays), jazz, R\&B, etc. (music genre of Grant Green)\\
    Context Object Possible Answers & Jazz. \\
    Context-Based Candidates & Bee, Beach \\
    \midrule
    Context-Influenced? & False \\
    Correct Class? & False\\
    Exists Answer that Satisfies Both? & False \\

    \midrule
    \multicolumn{2}{l}{\textbf{Example 7}} \\
    \midrule
    Context & \textbf{Samuil Marshak} passed away at \textbf{Moscow}. \\
    Query & Jean Metcalfe, who is employed by \\
    Class & Company/Person \\
    \midrule
    Context Subject Possible Answers &  Russia-1, Channel One Russia, RT, TV Rain, etc.\\
    Context Object Possible Answers &  Russia-1, Channel One Russia, RT, TV Rain, etc.\\
    Context-Based Candidates & BBC, Radio \\
    \midrule
    Context-Influenced? & False \\
    Correct Class? & True\\
    Exists Answer that Satisfies Both? & False \\
    
    \bottomrule
    \bottomrule

\label{tab:annotation_examples}
    
\end{longtable}

\end{document}